\documentclass[11pt]{article}

\usepackage[preprint]{acl}

\usepackage{times}
\usepackage{latexsym}

\usepackage[T1]{fontenc}

\usepackage[utf8]{inputenc}

\usepackage{microtype}

\usepackage{inconsolata}
\usepackage{booktabs}

\usepackage{graphicx}
\usepackage{afterpage}
\usepackage{amsmath}
\usepackage{hyperref} 
\usepackage{cleveref} 

\definecolor{lightgray1}{gray}{0.9}
\definecolor{lightgray2}{gray}{0.7}

\usepackage{graphicx}
\usepackage{amssymb}
\usepackage{bbm}
\usepackage{placeins}
\usepackage{enumitem}
\usepackage{subcaption}
\usepackage{color}
\usepackage{multirow}
\usepackage{tabularx}
\usepackage{bm}
\usepackage{xcolor}
\definecolor{arxivblue}{HTML}{0077b6}
\usepackage[normalem]{ulem}
\usepackage{soul}
\usepackage{tcolorbox} 
\usepackage{algorithm2e}
\RestyleAlgo{ruled}

\usepackage{sidecap}
\usepackage{xcolor,colortbl}
\usepackage{float}
\usepackage{hyperref}
\usepackage{dblfloatfix}

\usepackage{algorithm2e}
\RestyleAlgo{ruled}

\usepackage{sidecap}
\usepackage{authblk}

\useunder{\uline}{\ul}{}
\newcommand{\myparagraph}[1]{\vspace{1.5pt}\noindent{\bf #1}}
\makeatletter
\newcommand*{\rom}[1]{\expandafter\@slowromancap\romannumeral #1@}
\makeatother

\title{More Images, More Problems? A Controlled Analysis of VLM Failure Modes.}

\author{Anurag Das$^{1}$\thanks{Work done during internship at Samsung AI, Cambridge.}, Adrian Bulat$^{2,3}$, Alberto Baldrati$^{2}$, Ioannis Maniadis Metaxas$^{2}$ \\ Bernt Schiele$^{1}$, Georgios Tzimiropoulos$^{2,4}$, Brais Martinez$^{2}$ \\
\small $^1$MPI for Informatics, Saarland Informatics Campus $^2$Samsung AI, Cambridge \\
\small $^{3}$Technical University of Iași, Romania $^4$Queen Mary University of London, UK \\
\small andas@mpi-inf.mpg.de
}

\begin{document}
\maketitle

\begin{abstract}
Large Vision Language Models (LVLMs) have demonstrated remarkable capabilities, yet their proficiency in understanding and reasoning over multiple images remains largely unexplored. While existing benchmarks have initiated the evaluation of multi-image models, a comprehensive analysis of their core weaknesses and their causes is still lacking. 
In this work, we introduce MIMIC (Multi-Image Model Insights and Challenges), a new benchmark designed to rigorously evaluate the multi-image capabilities of LVLMs. Using MIMIC, we conduct a series of diagnostic experiments that reveal pervasive issues: LVLMs often fail to aggregate information across images and struggle to track or attend to multiple concepts simultaneously. To address these failures, we propose two novel complementary remedies. On the data side, we present a procedural data-generation strategy that composes single-image annotations into rich, targeted multi-image training examples. On the optimization side, we analyze layer-wise attention patterns and derive an attention-masking scheme tailored for multi-image inputs. Experiments substantially improved cross-image aggregation, while also enhancing performance on existing multi-image benchmarks, outperforming prior state of the art across tasks. Data and code will be made available at \url{https://github.com/anurag-198/MIMIC}.

\end{abstract}    
\section{Introduction}
\label{sec:intro}

Current Large Vision Language Models \cite{li2024llava,liu2024improved,wang2024cogvlm,wang2024qwen2,yao2024minicpm} showcase impressive vision-language understanding capabilities~\cite{goyal2017making,mathew2021docvqa,kembhavi2016diagram}. Most of these models are built upon pre-trained vision encoders~\cite{radford2021learning,zhai2023sigmoid} and large language models (LLMs)~\cite{touvron2023llama,abdin2024phi,jiang2024mixtral}. While early efforts primarily focused on single images~\cite{liu2024improved}, recent works have extended them to support multiple images~\cite{li2024llava,wang2024qwen2,chen2023minigpt} and videos~\cite{zhang2023video} by incorporating temporal modeling and adjusting the positional embeddings~\cite{zhang2023video,li2024llava}.

Despite their success, LVLMs continue to face significant challenges\cite{stogiannidis2025mind,liu2023mitigating,ouali2024clip,guan2024hallusionbench,kaul2024throne,qian2024easy}. Progress towards identifying and addressing these challenges can follow two primary avenues: the development of comprehensive evaluation benchmarks~\cite{goyal2017making,mathew2021docvqa,kembhavi2016diagram,masry2022chartqa,li2024seed,fu2024mme} and the study of the models' inner workings~\cite{deng2025words,qian2024easy,kaul2024throne}. To date, research in both areas has predominantly focused on the single-image setting. While early efforts have introduced benchmarks for multi-image scenarios~\cite{wang2024muirbench,jiang2024mantis,fu2024blink}, a comprehensive, in-depth analysis to ascertain the true efficacy of these models and identify the root causes of their limitations is notably absent. 

In this work, we address this gap by conducting a systematic study of LVLMs in multi-image contexts. We first analyze and characterize common failure modes using a newly proposed benchmark, and then seek to mitigate these limitations using two novel complementary fine-tuning strategies. Our in-depth analysis is performed on the newly introduced \textbf{MIMIC} (\textbf{M}ulti-\textbf{I}mage \textbf{M}odel \textbf{I}nsights and \textbf{C}hallenges) benchmark. Built from MS-COCO~\cite{lin2014microsoft}, using its bounding boxes and class labels, MIMIC procedurally generates multi-image sequences by leveraging per-image annotations that give fine-grained control over information spread, distractor presence, object-instance distributions, sequence length, and query complexity, while providing unambiguous ground-truth answers for robust, decorrelated analysis of the model's strengths and weaknesses. Using both quantitative and qualitative assessments, our study reveals that current state-of-the-art LVLMs struggle to effectively aggregate information across multiple images, are unable to track/attend to multiple concepts simultaneously, while being susceptible to distractors. 
We attribute these shortcomings to a combination of factors, including limitations in multi-image sequence modeling, training data biases, poor inter-image communication induced by the causal attention and the inherent complexity of multi-image reasoning tasks. 

Finally, to address the identified problems, we propose two new finetuning strategies: (1) a data-centric approach that generates targeted multi-image training examples to provide rich, multi-image supervision derived from OpenImages~\cite{kuznetsova2020open}; and (2) an optimization-centric approach that leverages layer-wise attention analysis to derive an attention-masking scheme tailored for multi-image inputs. Our proposed finetuning strategies lead to substantial performance gains across all scenarios.

In summary, our main contributions are:

\begin{itemize}[noitemsep,topsep=0pt]
    \item We introduce MIMIC, a comprehensive evaluation framework for multi-image LVLMs that probes various aspects of model performance through a controlled and diverse set of tasks.
    \item We conduct an extensive evaluation of several state-of-the-art LVLMs using MIMIC, uncovering critical insights into their capabilities and limitations in multi-image settings.
    \item We propose a novel data-centric finetuning approach using synthetically generated multi-image data, alongside an optimization-centric attention-masking strategy, both of which significantly enhance model performance in multi-image contexts.
    \item  We set new state-of-the-art results on existing multi-image benchmarks, demonstrating the effectiveness of our proposed methods.

\end{itemize}

\begin{figure*}[ht!]
\centering
\includegraphics[width=\linewidth]{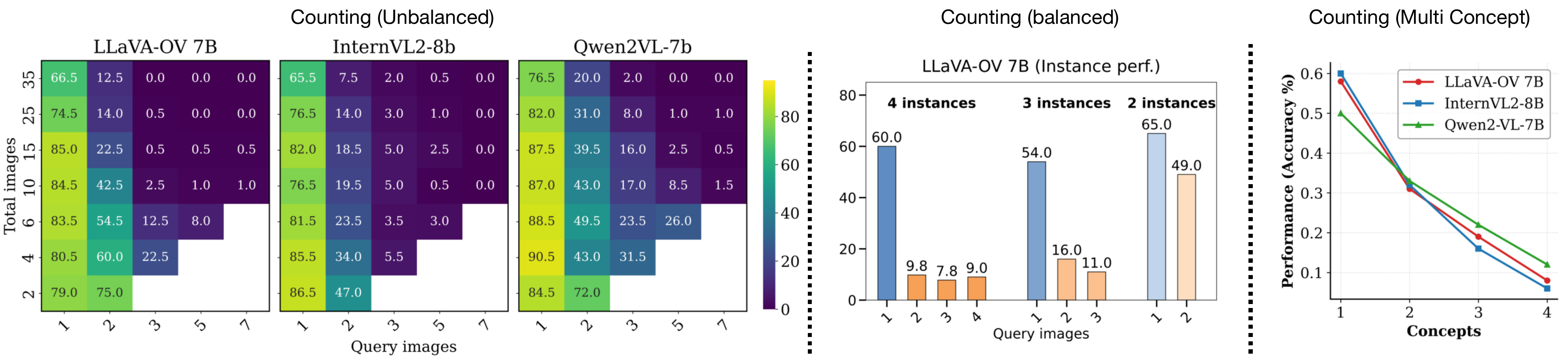}
\caption{
\textbf{Counting performance under different settings.}
\textbf{Left (Unbalanced):} We compare different LVLMs by analyzing the trade-off between the number of query images and the total number of images without controlling for number of instances.
\textbf{Mid (Balanced)}: We fix the total number of images to 7 and of object instances distributed across query images to 4,3 and 2. In both settings, performance consistently drops when instances are spread over multiple images. \textbf{Right (Multi Concept)}: We increase the complexity by adding more classes (concepts) to the counting task, and observe a steep performance drop, indicating limited capacity for multi-concept tracking.} 
\label{fig:query_total_image}
\vspace{-0.3cm}
\end{figure*}

\section{Related work}
\noindent \textbf{Multi-Image Large Vision Language Models:} 

Early LVLMs such as Flamingo~\cite{alayrac2022flamingo} and PaLM-E~\cite{driess2023palm} pioneered the integration of pre-trained vision encoders with powerful LLMs for VQA and captioning. Subsequent models~\cite{dai2023instructblip,li2024llava} introduced expanded instruction tuning and multi-modal pre-training techniques. More recent advancements include MiniGPT-5~\cite{zheng2023minigpt}, Qwen2-VL~\cite{wang2024qwen2}, CogVLM2~\cite{hong2024cogvlm2} and InternVL3~\cite{zhu2025internvl3} further advanced the field by scaling training data and model capacity and adopting more sophisticated architectural designs. While early LVLMs primarily operated on low-resolution, single-image inputs~\cite{liu2024improved,lin2024vila}, later research significantly expanded their scope. High-resolution images~\cite{li2024llava,wang2024qwen2,zheng2023minigpt} are commonly processed by splitting them into fixed-resolution patches and treating them as image sequences, while videos are represented by extracting frames to form multi-image inputs. In addition, models have begun to explicitly support multi-image context~\cite{wang2024qwen2,li2024llava,zhu2025internvl3}, enabling reasoning across multiple visual inputs. Multi-image capability is introduced by fine-tuning single-image LVLMs on multi-image instruction-tuning data, while largely preserving the original model architecture and attention mechanisms.

\paragraph{Evaluation of LVLMs:} Early evaluation efforts focused on narrower domains with benchmarks such as MS-COCO~\cite{lin2014microsoft}, VQA~\cite{antol2015vqa}, DocVQA~\cite{mathew2021docvqa}, GQA~\cite{hudson2019gqa} and AI2D~\cite{kembhavi2016diagram}, primarily assessing single-image understanding and using templetized questions with limited diversity. Later work introduced more comprehensive benchmarks to evaluate a wider range of skills, e.g. SEED-Bench~\cite{li2024seed}, MMBench~\cite{liu2024mmbench} and MME~\cite{fu2024mme}, which feature diverse question types and require complex reasoning abilities. Similarly, video benchmarks such as MMVU~\cite{zhao2025mmvu} and VideoMME~\cite{fu2025video} require models to understand temporal dynamics and to reason across multiple frames.

Closer to our work, several benchmarks have been proposed specifically for multi-image LVLMs. MuirBench~\cite{wang2024muirbench} introduced 12 tasks evaluating multi-image understanding, including image comparison and multi-image reasoning. Blink~\cite{fu2024blink} included 14 tasks deemed easy for humans, highlighting LVLMs limitations in truly understanding multi-image visual content. Visual Haystack~\cite{wu2024visual} focuses on retrieval-based tasks, assessing how well models can find certain concepts within a long sequence of images. Instead, we provide a more granular analysis of model performance across various controlled dimensions, such as information distribution, query complexity and distractor presence. Moreover, while prior works often repurposed existing datasets, we design our task from scratch to allow for selective performance exploration. This allows us to pinpoint specific strengths and weaknesses in current models that prior benchmarks may have overlooked and, importantly, provide deeper actionable insights on their underlying root causes.

\paragraph{Analysis of LVLMs:} Parallel to the development of benchmarks, there is growing interest in analyzing the internal mechanisms of LVLMs to better root-cause their limitations at a data and architecture level. 
Current studies have investigated issues such as hallucination~\cite{liu2023mitigating,ouali2024clip,guan2024hallusionbench}, 
modality bias~\cite{deng2025words,wang2025text}, and sensitivity to input phrasing~\cite{qian2024easy}. These works often involve probing the models with carefully-designed inputs to reveal their decision-making process. 
Only recently have such analyses expanded to multi-image LVLMs~\cite{wang2024longllava,wu2024visual,sharma2024losing}. The closest to our work is the study by Wu et al.~\cite{wu2024visual}, which examines the retrieval capabilities of multi-image LVLMs as the sequence length increases, showing limitations when operating over long sequences. However, their focus is primarily on the models' ability to locate specific items within an image set and does not control conflating factors, nor seek to identify the root causes beyond data scarcity.

Instead, we systematically probe additional dimensions of multi-image understanding, such as information aggregation and multi-concept tracking. To control for confounding factors, our evaluation is designed to isolate specific unitary aspects of multi-image understanding, leading to precise conclusions and to the identification of areas for improvement. Moreover, we analyze the internal model's behavior, and complement our analysis with proposed solutions to address the identified challenges at both data and optimization levels.

\section{Challenges and Insights in Multi-Image LVLMs} \label{sec:challenges_and_insights}

Herein, we systematically investigate the current LVLMs limitations in multi-image scenarios across six complementary dimensions: information distribution, query complexity, reasoning patterns, robustness to visual distractors, scalability with the number of images, and multi-concept tracking. For this purpose, we introduce MIMIC, a controlled testbed synthesized from a curated subset of MS-COCO~\cite{lin2014microsoft}. Using the manually annotated bounding boxes and labels, MIMIC generates multi-image sequences that allow precise control over information spread, distractor presence, object-instance distributions, and sequence length. This design enables decorrelated, fine-grained analyses of the model's behavior. Beyond these dimensions, our framework scrutinizes the models' mechanisms for aggregating and reasoning over distributed visual information. Through this controlled analysis, we aim to isolate the specific limitations and offer actionable insights for the next generation of visual understanding models.

Section~\ref{ssec:tested} describes the tasks design and dataset construction of our probing benchmark.
Section~\ref{ssec:empirical} describes the systematic probing of several state-of-the-art LVLMs to uncover their strengths and limitations in multi-image understanding.

\begin{figure}
\centering
\includegraphics[width=\linewidth]{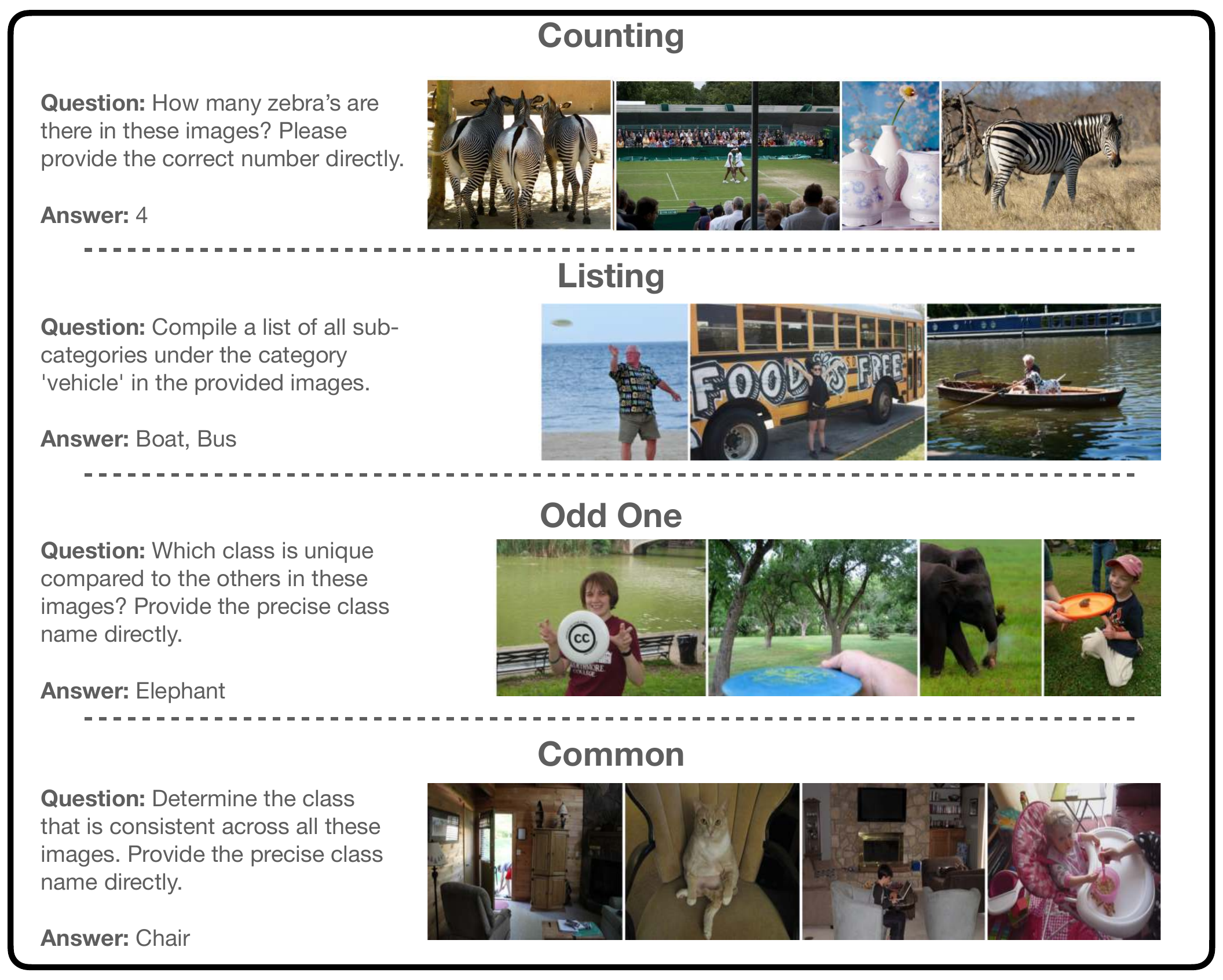}
\caption{\textbf{MIMIC Bench}: examples of each task.}
\vspace{-0.4cm}
\label{fig:mimic_bench_qual}
\end{figure}

{
 \setlength{\tabcolsep}{1pt}
 \renewcommand{\arraystretch}{1.0}
\begin{table}[ht!]
 \centering
\resizebox{\linewidth}{!}{%
        \begin{tabular}{l c c c c c c }
            \toprule
             &  \multicolumn{2}{c}{Counting} & \multirow{2}{4em}{\centering Common} &  \multirow{2}{4em}{\centering Odd One} &  \multirow{2}{4em}{\centering Listing} &  \multirow{2}{4em}{\centering Overall}  \\
             \cmidrule{2-3}
            & Bal. & Unbal. &  &  &  \\
            \midrule
            Queries & 5000 & 5800 & 1000 & 1000 & 1000 & 13800 \\
            Images & 5370 & 3761  & 3842 & 3726 & 4137 & 13145 \\
            Obj. inst. per query & 44.3 & 132.1 & 26.1 & 20.7 & 27.9 & 77.0  \\
            Min img per query  & 7 & 2  & 3 & 4  & 2 & 2\\
            Max img per query & 7 & 35 & 8 & 6 & 8 & 35 \\
            Median img per query & 7.0 & 15.0  &  5.0 & 5.0 & 5.0 & 7\\
            Avg words per question & 15.1 & 15.2 & 14.7 & 17.3 & 13.6 & 15.2 \\
            \bottomrule 
         \end{tabular} 
}
\caption{MIMIC Benchmark statistics per task. Counting settings: Balanced (Bal) and Unbalanced (Unbal).}
\vspace{-0.4cm}
\label{tab:our_dataset_distribution}
\end{table}
}
\subsection{Testbed benchmark construction}\label{ssec:tested}

We build the probing dataset by procedurally generating multi-image, open-ended question-answering tasks that target distinct aspects of cross-image reasoning. To this end, we sample a curated subset from MS-COCO~\cite{lin2014microsoft} by filtering images with object bounding boxes less than 5\% of the image in order to ensure visual recognizability at common LVLM input resolutions (e.g. LLaVA-OV's $384 \times 384$px). 
To minimize the impact of potential class imbalance, we first select a pool of classes and then sample from this pool, ensuring that each class is chosen with an equal probability. This ensures that the distributions of classes and instances remain consistent across settings.

MIMIC defines four core tasks: Counting, Listing, Common, and Odd-One, each targeting a distinct facet of multi-image reasoning. Fig.~\ref{fig:mimic_bench_qual} provides some qualitative examples, and Table~\ref{tab:our_dataset_distribution} reports dataset statistics. All tasks are formulated as open-ended question answering rather than multiple-choice to increase challenge, avoid shortcuts introduced by fixed option sets, and remove the need to calibrate distractor choices. To further reduce prompt bias, we employ multiple templatized prompts per task (see appendix for a template list). Below, we describe each task in detail.

\begin{figure}[ht!]
\centering
\includegraphics[trim=10pt 5pt 5pt 6pt, clip,width=\linewidth]{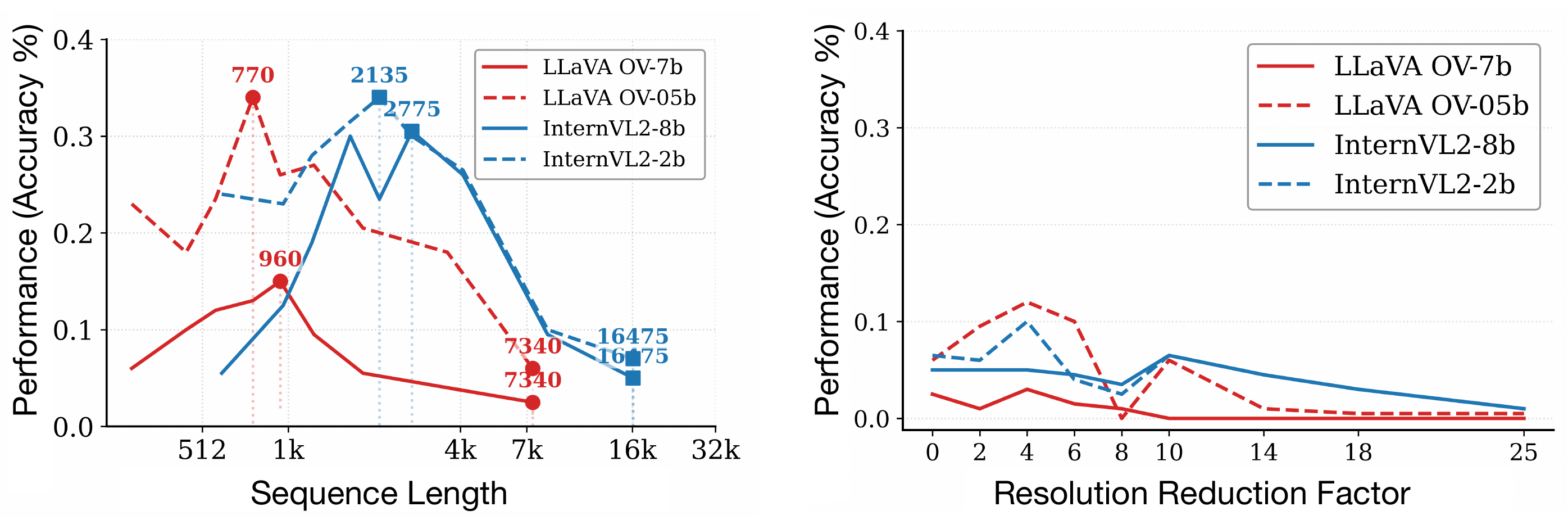}
\caption{\textbf{Effect of vision token sequence length on performance.} \textbf{Left}: Sequence length reduction via 1-D pooling. The square denotes the original sequence length. \textbf{Right}: Control experiment reducing the information via pixel space pooling while keeping the sequence length fixed. Results are reported for counting task with 3 query images and total 10 images.
}

\label{fig:seq_length_1d_pooling}
\end{figure}

\noindent \textbf{Counting}: Given a set of $N$ input images, and a query containing $k$ object classes, the model is asked to count the total number of instances of each class. With increasing difficulty, we vary the distribution of object instances across images. 
For example, in the easiest setting, all may be concentrated in a single image, while in more challenging cases, instances are spread across multiple images. We refer to this as the \textit{information spread}. Additionally, we introduce distractors - images that do not contain any instances of the target objects - to assess the model's ability to focus on relevant information. In summary, this task offers the following controllable dimensions: (1) number of object classes to count ($k$); (2) information spread across images ($s$); (3) number of distractor images and (4) total number of images. Each case probes different aspects and potential biases. For instance, increasing the number of object classes $k$ tests the model's multi-concept tracking ability, while varying the information spread evaluates its capacity to aggregate information across images.

To account for potential biases caused by the long-tail distribution of object instance counts in natural images, which may lead to models favoring smaller counts, we design two distinct settings: (1) \textit{Balanced}, where the total number of object instances is fixed, but distributed across a varying number of images; (2) \textit{Unbalanced}, where the total number of object instances varies arbitrarily with the number of images. The metric of choice is binary accuracy, i.e. the answer is correct if it matches the ground truth count exactly.

\noindent \textbf{Listing}: The model is presented with a set of $N$ images and asked to list all object classes belonging to a given category (e.g.: animals, vehicles, etc.) that it can identify. This task evaluates the model's ability to exhaustively extract information in a dense manner from multiple images. As a byproduct, it also measures the model's visual perception ability to recognize and categorize multiple objects, as well as its capacity to aggregate this information into a coherent list.
Similar to the Counting task, we vary the number of images and the distribution of object instances to assess the model's robustness in multi-image understanding.
The model's response is evaluated on the completeness and accuracy of the list, using F1-score as metric. 
See the appendix for a complete hierarchy of object categories and subcategories.

\noindent \textbf{Common and Odd-One}: The two tasks are designed to assess the model's ability to identify shared or unique elements across multiple images. Importantly, while previous tasks focus on aggregating information, these tasks require comparative reasoning across images, hence the model must first implicitly identify all objects before performing cross-image analysis.
In the Common task, the model has to determine which object class is present in all provided images, while in the Odd-One case, it must identify the object class that is present in a minority of images. For simplicity, we ensure by design that the answers are unique. The model's answers are evaluated based on their correctness, with binary accuracy as the metric.

\subsection{Empirical analysis}\label{ssec:empirical}

\noindent \textbf{Setup:} We evaluate several state-of-the-art LVLMs: LLaVA-OV~\cite{li2024llava}, Qwen2-VL~\cite{wang2024qwen2} and InternVL2~\cite{chen2024internvl}. We use publicly available checkpoints and follow the official data processing pipeline. For test data, we use the MIMIC benchmark described in Section~\ref{ssec:tested}, selecting tasks and configurations that best isolate the dimensions we aim to probe.

\noindent \textbf{Performance in Relation to Sequence Length and Number of Images:} LLMs are known to manifest position and sequence length biases~\cite{ravaut2024context}, with tokens appearing earlier and late in the sequence receiving more attention. 
Unlike for LLMs, we distinguish two axes of sequence length growth: (1) increasing the number of images, and (2) increasing the input image(s) resolution. We seek to understand if the performance degradation stems from the model's inability to handle long sequences, or from the inability to process many images. We disentangle these two factors with the following experiments:
(a) directly increasing the number of images without explicitly controlling for sequence length. In this setting, we simply vary the number of images provided to the model, measuring performance on the counting task. As the results from Fig.~\ref{fig:query_total_image} (left) show, performance degrades in all setting consistently for all models as the total number of images increases from 2 to 35s.
(b) reducing the vision token sequence length through 1-D average pooling applied to the original multi-image vision tokens. To ensure that the observed behavior is not an artifact of reduced information, we also perform a control experiment where we similarly reduce the amount of information by downsampling and then upsampling back the images in pixel space, prior to being passed to the vision encoder. This preserves the initial sequence length but reduces the amount of visual information available to the model.
This allows us to assess if the performance degradation observed in (a) is primarily due to the increased sequence length or number of images. 

The results are summarized in Fig.~\ref{fig:seq_length_1d_pooling}. On the left, we plot performance changes for different models as we decrease the number of vision tokens via 1-D pooling. Due to different processing, each model allocates different number of tokens per image, hence we mark two points - extreme right (no downsampling) and central point that maximizes performance. 
On the right, we show the control experiment, that decreases the information in the pixel space artificially without reducing sequence length.
Surprisingly, we find that reducing the sequence length in a zero-shot manner via 1-D pooling up to $4-8\times$ leads to significant performance improvements across all models. The control experiment confirms that gains are due to sequence length reduction rather than \textit{information reduction}.
This suggests that the models primarily struggle with long sequence understanding rather than with processing multiple distinct images.

\begin{tcolorbox}[
    colback=arxivblue!8!white, 
    colframe=arxivblue!40!cyan, 
    fonttitle=\bfseries,
    fontupper=\fontsize{11}{13}\selectfont, 
    boxsep=5pt,
    left=2pt,
    right=2pt,
    top=2pt,
    bottom=2pt,
    arc=2mm, 
    outer arc=2mm
]
\textbf{Finding 1:} The performance degradation in multi-image scenarios stems primarily from \textbf{increased sequence length} rather than the increased number of images.
\end{tcolorbox}
\begin{figure*}[ht!]
\centering
\includegraphics[width=\textwidth]{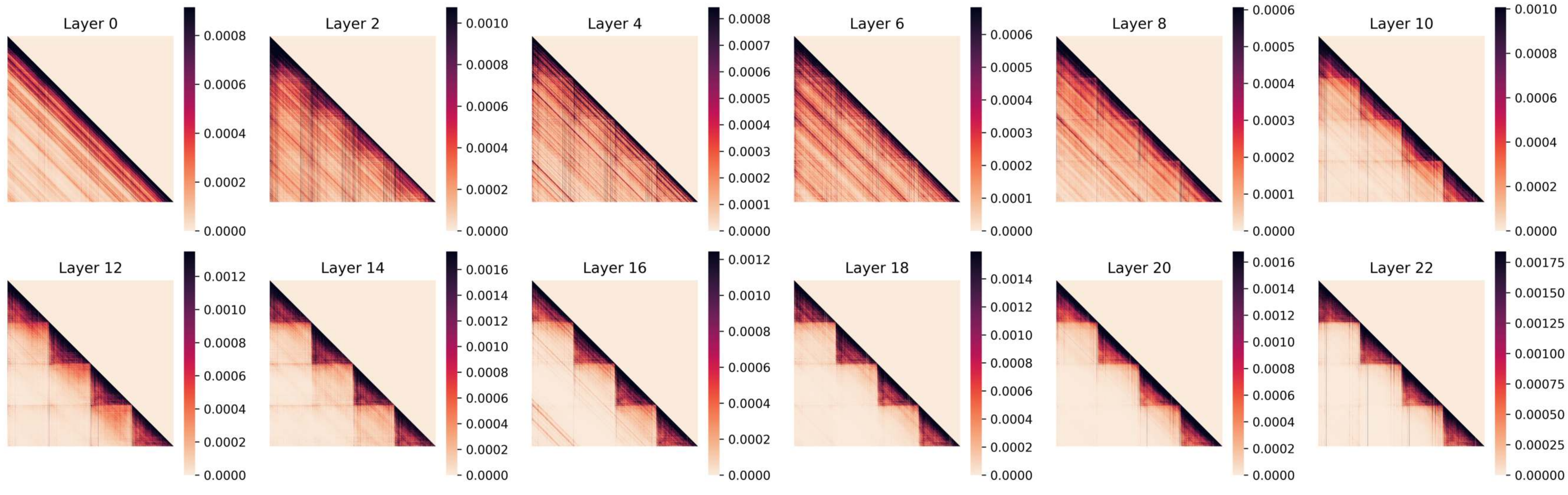}
\caption{\textbf{Inter-image and intra-image token attention across layers.} The attention patterns transitions from cross-image to intra-image interactions as we advance in depth.}
\label{fig:im2imattn}
\end{figure*}

Moreover, we observe that for LLaVA-OV performance peaks when the vision sequence length is approximately that of one or two images (i.e., roughly the number of vision tokens for a $384\times384$ image/patch). This suggests the model effectively relies on a single-image context and has limited practical multi-image integration; we later evaluate how targeted finetuning can mitigate this limitation.

\begin{tcolorbox}[
    colback=arxivblue!8!white, 
    colframe=arxivblue!40!cyan, 
    fonttitle=\bfseries,
    fontupper=\fontsize{11}{13}\selectfont, 
    boxsep=5pt,
    left=2pt,
    right=2pt,
    top=2pt,
    bottom=2pt,
    arc=2mm, 
    outer arc=2mm
]
\textbf{Finding 2:} Current LVLMs primarily \textbf{behave as single-image models}: performance peaks when the vision-token sequence length matches that produced by one or two images.
\end{tcolorbox}

\noindent \textbf{Information aggregation across images:} Prior benchmarks rarely control for how information is distributed across images, making it difficult to isolate whether models can effectively aggregate information across images. To this end, we vary the \textit{information spread} in the counting task, which defines how object instances are distributed across images. In Fig.~\ref{fig:query_total_image} (left and middle) we show the results of increasing the number of images containing the object instance from 1 to 7. We observe a sharp accuracy drop that approaches 0 even when very few distractors are present. This trend is consistent across all models tested and manifests both in balanced and unbalanced counting settings. 
This indicates that the models may rely on shortcuts, such as focusing on a single or very small subset of images, rather than effectively integrating information from all provided images.

\begin{tcolorbox}[
    colback=arxivblue!8!white, 
    colframe=arxivblue!40!cyan, 
    fonttitle=\bfseries,
    fontupper=\fontsize{11}{13}\selectfont, 
    boxsep=5pt,
    left=2pt,
    right=2pt,
    top=2pt,
    bottom=2pt,
    arc=2mm, 
    outer arc=2mm
]
\textbf{Finding 3:} Current LVLMs \textbf{struggle to aggregate information} across multiple images.
\vspace{-5pt}
\end{tcolorbox}

\noindent \textbf{Robustness to visual distractors.} In real-world scenarios, models often encounter irrelevant or distracting information. To evaluate the robustness of LVLMs, we introduce a varying numbers of irrelevant images into the input sequence. As shown in Fig.~\ref{fig:query_total_image} (left), the accuracy decreases as the number of distractors increases (e.g: from 79.0\% to 66.5\% (1 vs 34 distractors) for 1 query image, from 75.0\% to 12.5\% for two query images, etc., for LLaVA-OV). A particularly pronounced drop occurs as the number of images containing the object of interest increases, suggesting that distractors exacerbate the models' existing difficulties in aggregating information across multiple images.

\begin{tcolorbox}[
    colback=arxivblue!8!white, 
    colframe=arxivblue!40!cyan, 
    fonttitle=\bfseries,
    fontupper=\fontsize{11}{13}\selectfont, 
    boxsep=3pt,
    left=2pt,
    right=2pt,
    top=2pt,
    bottom=4pt,
    arc=2mm, 
    outer arc=2mm
]
\textbf{Finding 4:} Models \textbf{are sensitive to visual distractors}, especially if information is spread out.
\end{tcolorbox}

\noindent \textbf{Multi-concept tracking.} The ability to track and attend to multiple concepts simultaneously is critical for multi-image understanding. To probe this capability, we vary the number of object classes $k$ that the model is required to count. As shown in Fig.~\ref{fig:query_total_image} (right), the model performance degrades sharply as $k$ increases, indicating a limited capacity to handle multiple concepts at once. 

\begin{tcolorbox}[
    colback=arxivblue!8!white, 
    colframe=arxivblue!40!cyan, 
    fonttitle=\bfseries,
    fontupper=\fontsize{11}{13}\selectfont, 
    boxsep=4pt,
    left=2pt,
    right=2pt,
    top=2pt,
    bottom=4pt,
    arc=2mm, 
    outer arc=2mm
]
\textbf{Finding 5:} LVLMs demonstrate \textbf{limited capacity for multi-concept tracking}, reducing their reliability on complex multi-object queries.
\end{tcolorbox}

{
 \setlength{\tabcolsep}{2pt}
 \renewcommand{\arraystretch}{1.0}
\begin{table*}[ht!]

 \centering
 \resizebox{\linewidth}{!}{%
\begin{tabular}{l|c|c|c|c|c|c |c|c|c|c|c|c|c }
\toprule
Model  & Geographic. & Counting & Action. &  Grounding & Matching. & Ordering & Scene. & Difference. & Cartoon. & Diagram & Attribute. & Retrieval & Overall  \\
 \hline 
Random Choice & 25.0 & 21.0 & 23.4 & 25.0 &  24.1 & 22.8 & 25.0 &  23.2 & 25.0 &  29.6 &  20.0 & 21.3 & 24.0\\
 Human & 98.0 & 94.9  & 97.6 & 85.7 & 94.8 & 87.5 & 94.6 & 92.9 & 82.1 & 98.99 &  87.6 & 86.3 & 93.2\\
\hline
GPT-4o~\cite{openai2023gpt} & 56.0 & 49.2 & 44.5 & 36.9 & 86.9 & 23.44 & 71.5 & 60.3 & 51.3 & 88.7 & 56.1 & 80.1 & 68.0 \\
Gemini Pro~\cite{team2023gemini} & 48.0  & 28.6 & 36.0 & 28.6 & 66.6 & 12.5 & 59.1 & 45.3 & 47.4 & 64.8 & 41.3 & 43.8 &  49.4 \\

Mantis-8b-Idefics2~\cite{jiang2024mantis} & 26.0 & 38.5 & 33.5 & 26.2 & 53.9 &  18.8 &  57.0 & 28.8 & 38.5 & 67.6  & 48.5 & 35.6 &  44.5\\

Idefics-9B-Instruct~\cite{laurenccon2023obelics} & 35.0 & 21.8 & 26.2 & 26.2 & 24.8 & 15.6 &  56.5 & 27.6 & 39.7  & 25.4 & 17.9 & 17.1 & 35.4 \\

Emu2-Chat(37B)~\cite{sun2024generative} & 34.0 & 31.2 & 27.4 & 26.2 & 37.3 & 15.6 & 48.4 & 32.6 & 43.6 & 37.7 & 31.6 & 24.0 &  33.6 \\

VILA1.5-13B~\cite{lin2024vila} & 31.0 & 19.7 & 28.7 & 25.0 & 40.9 & 10.9 & 56.5 & 24.7 & 30.8 & 42.7 & 24.5 & 30.1 & 33.1\\


LLaVA-NeXT-34B~\cite{liu2024llava} & 12.0 & 36.3 & 26.2 & 33.3 & 37.9 & 21.9 & 54.3 & 22.1 & 41.0 & 38.2 & 38.3 & 25.0 & 33.3\\

 LLaVA-v1.5-7B~\cite{liu2024improved} & 20.0 & 23.1 & 27.4 & 14.3 &  23.5 & 23.4 & 35.0 & 20.0 & 24.4 & 25.1 & 23.0 & 19.9 & 23.5 \\

 LLaVA-v1.5-13B~\cite{liu2024improved} & 20.0 & 25.2 & 29.3 & 14.3 & 20.3 & 20.3 & 36.6 & 20.0 & 25.6 & 31.7 & 23.0 &  20.9 & 24.4 \\

CogVLM~\cite{wang2024cogvlm} & 13.0 & 14.1 & 26.2 & 16.7 & 21.3 & 12.5 & 41.4 & 19.7 & 41.0 & 19.6  & 16.3 &  15.8 & 20.9\\

MiniGPT-4-v2~\cite{chen2023minigpt} & 13.0 & 12.0 & 14.0 & 25.0 & 17.0 & 18.8 & 14.5 & 20.0 & 21.8 & 21.6 & 17.4 & 14.7 & 17.4\\

Qwen2-VL-7B~\cite{wang2024qwen2} & 12.0 & 38.9 & 42.7 & 28.5 & 57.5 & 10.9 & 75.3 & 32.9 & 38.5 & 49.2 & 46.4 & 26.7 & 43.0 \\
Qwen2-VL-2B~\cite{wang2024qwen2} & 14.0 & 27.8 & 35.4 & 26.2 & 34.3 & 10.9 & 51.1 & 19.4 & 39.7 & 21.4 & 31.1 & 15.4 & 27.2 \\

InternVL2-8B~\cite{chen2024expanding} & 17.0 & 30.3 & 34.7 & 28.5 & 43.5 & 17.2 & 60.2 & 26.2 & 46.2 & 46.5 & 42.8 & 33.6 & 37.9 \\
InternVL2-2B~\cite{chen2024expanding} & 17.0 & 21.8 & 26.8 & 26.2 & 31.7 & 10.9 & 52.2 & 17.4 & 35.9 & 21.6 & 16.8 & 13.7 & 24.3 \\

\hline 

LLaVA-OV-0.5B~\cite{li2024llava} & 22.0 & 20.9  & 31.1 & 25.0 & 30.2 & 7.8 & 46.7 & 24.1 & 42.3 & 25.1 & 23.9 &  20.5 & 26.8 \\

\rowcolor{lightgray1} Ours & 43.0 & 20.5 & 41.4 & 22.6 & 38.4 & 9.4 & 52.7 & 22.1 & 37.2 & 39.4 & 29.6 & 32.5 & 33.6 \\

 \rowcolor{lightgray1} Ours (Masked) & 31.0 & 17.5 & 40.8 & 33.3 & 37.3 & 14.1 & 53.8 & 21.5 & 42.3 & 34.9 & 28.1 & 32.9 & 32.5 \\

\hline 

LLaVA-OV-7B~\cite{li2024llava} & 43.3 & 24.8 & 36.6  & 29.7 & 45.3 & 17.2 & 71.5 & 30.0 & 35.9 & 54.2 & 32.7 & 46.2 & 41.7 \\
\rowcolor{lightgray1} Ours (Masked) & 44.0 & 35.9 & 51.2 & 42.9 & 59.9 & 12.5 & 71.0 & 43.5 & 38.5 & 62.1 & 52.0 & 48.6 & 51.3 \\

\bottomrule
\end{tabular}
}
\caption{Performance comparison across different MuirBench~\cite{wang2024muirbench} subtasks. \textbf{Ours (Masked)}: Our efficient model trained with LoRA and masked attention. \textbf{Ours}: Fully fine-tuned model. \textit{Due to computational constraints, we do not fully finetune LLaVA-7B model.}}

\label{tab:muirbench}
\end{table*}
}
\noindent \textbf{Multi-image interaction.} To probe how visual information is propagated and integrated across images at the token level, we analyze attention patterns among vision tokens in multi-image inputs.
Concretely, we compute the normalized attention scores from each vision token to all other vision tokens in the input sequence subject to an autoregressive attention masking on a subset of 50 samples. Fig.~\ref{fig:im2imattn} summarizes the results across a few layers of interest for a LLaVA-OV model for multi-image inputs with 4 images.
We find that in earlier layers, there is a significant amount of inter-image attention, indicating that the model is attempting to integrate information across images. However, as we progress to deeper layers, the attention becomes predominantly intra-image. This inflection point occurs somewhere around the middle of the network. 
This shift may contribute to the observed difficulty in aggregating information across multiple images. Conceptually, the build-up of representations appears to proceed from broad semantic correlations across images to finer-grained, instance-level integration.

This has a series of consequences: (1) the early inter-image attention may introduce noise or distractions that hinder the model's ability to focus on relevant information in later layers; hence, early mistakes in cross-image interaction are harder to correct; (2) the cross-image interaction under a causal attention mechanism may lead to error propagation, where tokens belonging to later images cumulate higher amount of noise with incorrect information from earlier images; this may reduce the vision perception capability of the model for later images and explain some of the performance degradation as the number of images increases; (3) the architecture and training objectives may not sufficiently encourage cross-image integration, leading to a default behavior of treating images independently; (4) the observed attention patterns may reflect inherent biases in the training data, where the multi-image tasks don't require deep cross-image reasoning, leading the model to learn shortcuts that prioritize single-image understanding.

\begin{tcolorbox}[
    colback=arxivblue!8!white, 
    colframe=arxivblue!40!cyan, 
    fonttitle=\bfseries,
    fontupper=\fontsize{11}{13}\selectfont, 
    boxsep=1pt,
    left=2pt,
    right=2pt,
    top=2pt,
    bottom=2pt,
    arc=2mm, 
    outer arc=2mm
]
\textbf{Finding 6:} \textbf{Inter-image attention diminishes} in deeper layers of LVLMs, indicating a shift from cross-image integration to intra-image focus.
\end{tcolorbox}

\section{Method}

In the previous section, we identified key limitations of LVLMs on multi-image tasks via zero-shot evaluation using the MIMIC benchmark. Here, we investigate targeted fine-tuning strategies derived from our findings and aimed at improving multi-image reasoning capabilities. In particular, we explore two complementary approaches: a data-centric fine-tuning strategy using synthetically generated multi-image data, and an optimization-centric attention-masking strategy.

\myparagraph{Multi-Image Finetuning:} We fine-tune LLaVA-OV models on a unified training dataset composed of samples procedurally generated using the MIMIC pipeline (see~\cref{ssec:tested}) together with the original LLaVA-OV multi-image instruction-tuning data (approximately 580K samples). Unlike the MIMIC benchmark used for evaluation, our fine-tuning data is built from OpenImages and provides explicit supervision for cross-image reasoning. It contains approximately 198K samples, with sequence lengths of up to 10 images (see appendix), deliberately exposing models to substantially longer vision-token sequences. All four MIMIC tasks are included to encourage diverse multi-image reasoning behaviors.

\myparagraph{Attention Masking:} Our analysis shows that inter-image attention diminishes in deeper layers (see~\cref{fig:im2imattn}). Motivated by this, we apply layer-wise attention masking during fine-tuning, restricting vision tokens to attend only to tokens from the same image in selected layers, while leaving text-token attention unchanged. This design offers two key benefits. First, it reduces unnecessary cross-image interactions, leading to a more efficient model with lower computational cost (see~\cref{tab:flops} and ~\cref{fig:masked_attn} of appendix). Second, it encourages cleaner image-local representations in deeper layers, which empirically improves performance across several benchmarks. For this setting, we employ LoRA-based fine-tuning to further improve parameter efficiency. See appendix for implementation details.

\section{Results}

\subsection{Comparison with state-of-the-art}

{
 \setlength{\tabcolsep}{2pt}
 \renewcommand{\arraystretch}{1.0}
\begin{table}
 \centering
\resizebox{\linewidth}{!}{%
        \begin{tabular}{l |c|c |c | c |c |c |c }
            \toprule
            Model & MuirBench  & Blink & MMIU  &  MIRB & MMT (val) & NLVR2 & Avg. \\
            \hline
            GPT-4V & 62.3 & 54.6 & - & 53.1 & 64.3 & - & - \\
            InternVL2-Llama3-76B 
            & 51.2 & 56.8 & 44.2 & 58.2 & 67.4 & - & -\\
            LLaVA-v1.5-7B & 20.0 & 37.1 & 19.2 & 28.5 & - & - & - \\  
            InternVL2-2B
            & 24.3 & 16.3 & 13.6 & 25.0 & 46.7 & 18.9 & 24.1\\
            InternVL2-8B
            & 37.9 & 23.4 & 36.8 & 48.6 & 57.9 & 8.7 & 35.6 \\
            Qwen2VL-2B
            & 27.2 & 12.7 & 38.7 & 45.9 &  51.9 & 41.6 & 36.3\\
            Qwen2VL-7B
            & 43.0 & 17.7 & 52.6 & 60.8 & 61.7 & 41.5 & 46.2 \\ 
            \hline
             LLaVA-OV-0.5B
             & 26.8  & 40.4 & 34.2 & 31.8 & 41.1 & 61.2 & 39.3 \\
          
            \rowcolor{lightgray1} Ours & 33.6  & 38.9 & 37.2 & 32.8 & 45.6 & 68.0 & 42.7\\
          
            \rowcolor{lightgray1} Ours (masked) & 32.5 & 39.1 & 36.3 & 28.5 & 45.9 & 65.1 & 41.2 \\
             
             \hline 
   
             LLaVA-OV-7B
             & 41.7  & 50.4 & 45.0 & 47.2 & 56.6 & 84.2 & 54.2 \\
    
            \rowcolor{lightgray1} Ours (masked) & 51.3 & 51.9 & 45.5 & 51.0 & 55.3 & 87.3 & 57.1\\
            \bottomrule 
         \end{tabular} 
}
\caption{
Comparisons on multi-image benchmarks: MuirBench, Blink, MMIU, MIRB, MMT, and NLVR2.
}
\label{tab:mmiu_blink}
\end{table}
}

{
 \setlength{\tabcolsep}{4pt}
 \renewcommand{\arraystretch}{1.0}
\begin{table}
 \centering
\resizebox{\linewidth}{!}{%
        \begin{tabular}{l|c|c|c|c|c}
            \toprule
            Model & Common &  Counting &  Odd-one & Listing & Avg.   \\
            \hline 
            Mantis-8B-llama3~\cite{jiang2024mantis} & 13.0 & 19.9 & 10.9 & 17.0 & 15.2\\
            InternVL2-2B~\cite{chen2024expanding}  & 25.6 & 11.7 & 9.6 & 19.6 & 16.6 \\
            InternVL2-8B~\cite{chen2024expanding}  & 45.2 & 18.9 & 30.2 & 29.8 & 31.0 \\
            Qwen2VL-2B~\cite{wang2024qwen2} & 41.9 & 21.7 & 30.2 & 23.8 & 29.4 \\
            Qwen2VL-7B~\cite{wang2024qwen2} & 58.6 & 35.7 & 35.9 & 23.4 &  38.4\\ 
            \hline
             LLaVA-OV-0.5B~\cite{li2024llava}  & 44.7 & 29.7 & 8.3 & 22.8 & 26.4  \\
         
             \rowcolor{lightgray1} Ours  & 68.5 & 37.8 & 41.0 & 34.5 & 45.5 \\
             \rowcolor{lightgray1} Ours (masked) & 68.9 & 35.8 & 50.9 & 42.0 & 49.4 \\
            
             \hline 
             LLaVA-OV-7B~\cite{li2024llava} & 71.5 & 29.7 & 58.1 & 56.6 & 54.0 \\ 
             \rowcolor{lightgray1} Ours (masked) & 75.5 & 51.2 & 72.1 & 55.0 & 63.8\\
            \bottomrule 
         \end{tabular} 
}
\caption{
Comparisons on our benchmark. We report model's accuracy for Odd-one, Common and Counting whereas f1 score for listing benchmark.}
\label{tab:mimic}
\end{table}
}

\begin{table*}
\centering
\small 
\setlength{\tabcolsep}{2pt} 
\renewcommand{\arraystretch}{1.2}

\begin{minipage}[b]{0.28\textwidth}
    \centering
    \resizebox{\textwidth}{!}{%
    \begin{tabular}{l|c|c|c|c}
    \toprule
     & Common & Count & Odd-one & List \\
    \hline
    \color{gray!80}Ours (all) & \color{gray!80}68.5 & \color{gray!80}37.8 & \color{gray!80}41.0 & \color{gray!80}34.5 \\
    LLaVA-OV-0.5B & 44.7 & 29.7 & 8.3 & 22.8 \\
    \hline  
    only Common & \color{gray!80} 73.7 & 32.0 & 3.7 & 30.7 \\
    only Counting & 35.8 & \color{gray!80} 39.4 & 12.2 & 20.7 \\
    only Odd One & 34.4 & 31.8 & \color{gray!80} 53.6 & 31.1 \\
    only Listing & 46.0 & 29.3 & 11.1 & \color{gray!80} 28.3 \\
    \bottomrule 
    \end{tabular}%
    }
\end{minipage}%
\hfill
\begin{minipage}[b]{0.24\textwidth}
    \centering
    \resizebox{\textwidth}{!}{%
    \begin{tabular}{l|c|c}
    \toprule
     & FLOPs (Gain) & Avg. Perf. \\
    \hline
    LLaVA-OV-0.5B & 58B (0\%) & 26.4\\
    Ours & 58B (0\%) & 45.5 \\
    Ours (masked) & 11.2B (81\%) & 49.4 \\
  
    \bottomrule 
    \end{tabular}%
    }
\end{minipage}%
\hfill
\begin{minipage}[b]{0.35\textwidth}
    \centering
    \resizebox{\textwidth}{!}{%
    \begin{tabular}{l|c|c|c|c|c}
    \toprule
    Layers masked & Comm. & Count. & Odd. & List. & Avg. \\ 
    \hline
    No mask. & 70.0 & 32.0 & 37.9 & 44.5 & 46.1 \\   
    0-23 & 64.5 & 36.1 & 20.9 & 29.2 & 37.7 \\
    0-11 & 62.5 & 27.3 & 28.8 & 33.6 & 38.1 \\
    \rowcolor{lightgray1} 12-23 & 68.9 & 35.8 & 50.9 & 42.0 & 49.4 \\
    \bottomrule
    \end{tabular}%
    }
\end{minipage}

\caption{\textbf{Left: }Cross-task generalisation. We train LLaVA-OV model individually with each task and compare cross task generalisation. Ours (with all task): upperbound trained with all 4 tasks. \textbf{Middle:} Efficiency Analysis of Masked Attention. \textbf{Right:} Ablation wrt different layers for attention
masking. Experiments on LLaVA-OV 0.5B.}
\label{tab:comb_ablation}
\end{table*}

\myparagraph{Existing multi-image benchmarks.}
We first report results on MuirBench~\cite{wang2024muirbench} and its subtasks in~\cref{tab:muirbench}. Across all model sizes, our approach consistently outperforms the corresponding LLaVA-OV baseline. Notably, for the 7B model, our masked-attention variant improves the overall score from 41.7 to 51.3\%. We observe a similar trend for the smaller 0.5B variant, indicating that the improvements are robust across model sizes. Interestingly, our method generalizes well to out-of-domain subtasks, including geographic, action and diagram understanding, suggesting that our data construction strategy teaches the model multi-image processing \textit{concepts} rather than \textit{object perception}, which we argue develops already in the single-image training phase.

Next, we extend the evaluation to additional multi-image benchmarks, including Blink~\cite{fu2024blink}, MMIU~\cite{meng2024mmiu}, MIRB~\cite{zhao2024benchmarking}, MMT~\cite{ying2024mmt}, and NLVR2~\cite{suhr2019corpus}. Our approach achieves consistent improvements across all variants. As shown in~\cref{tab:comb_ablation}, our masked-attention fine-tuning strategy yields significant gains over the baseline even with very few trainable parameters, and in some cases outperforms full fine-tuning (e.g., LLaVA-OV 0.5B).

\myparagraph{MIMIC benchmark.} We report results in~\cref{tab:mimic}. Unless otherwise specified, all results for Counting subtask correspond to the balanced split. Our method significantly outperforms LLaVA-OV across all four tasks. For the 0.5B model, the average score improves from 26.4 to 49.4, while for the 7B model, masked fine-tuning increases performance from 54.0 to 63.8. Gains are most pronounced on the Common and Odd-One tasks, highlighting improved information aggregation and multi-concept reasoning across images.

\subsection{Ablation studies and analysis}
\myparagraph{Cross-task generalization.} In this experiment, we train models on individual subtasks (e.g., Counting, Common, Odd-One, and Listing) to analyze their complementary roles and assess cross-task generalization. 
Table~\ref{tab:comb_ablation} (left) shows the results.
We observe that training on the Common task generalizes well to Counting and Listing, but not to Odd-One; a similar trend is observed when training on Odd-One. This behavior is expected, as the two tasks are complementary in nature: Common requires aggregating information across multiple images, whereas Odd-One emphasizes localizing distinctive evidence within a single image. Training on Listing consistently improves performance across all other tasks, while training on Counting primarily benefits Odd-One.

\myparagraph{Efficiency analysis.} 
Table~\ref{tab:comb_ablation} (mid) demonstrates that our masked attention variant achieves superior performance with substantially lower computational cost compared with vanilla attention. On the 0.5B backbone, masked finetuning reduces the FLOPs by $\sim$81\%, while outperforming full finetuning. This confirms that selectively constraining inter-image attention is both effective and efficient. See appendix for details on FLOPs estimations.

\myparagraph{Attention masking strategy.} 
Table~\ref{tab:comb_ablation} (right) ablates the layers at which attention masking is applied. Masking only deeper layers (layers 12–23) yields the best performance, whereas masking early layers significantly degrades accuracy. These results suggest that early layers are important for effective cross-image information aggregation.

\begin{figure}[ht!]
\centering
\includegraphics[width=\linewidth]{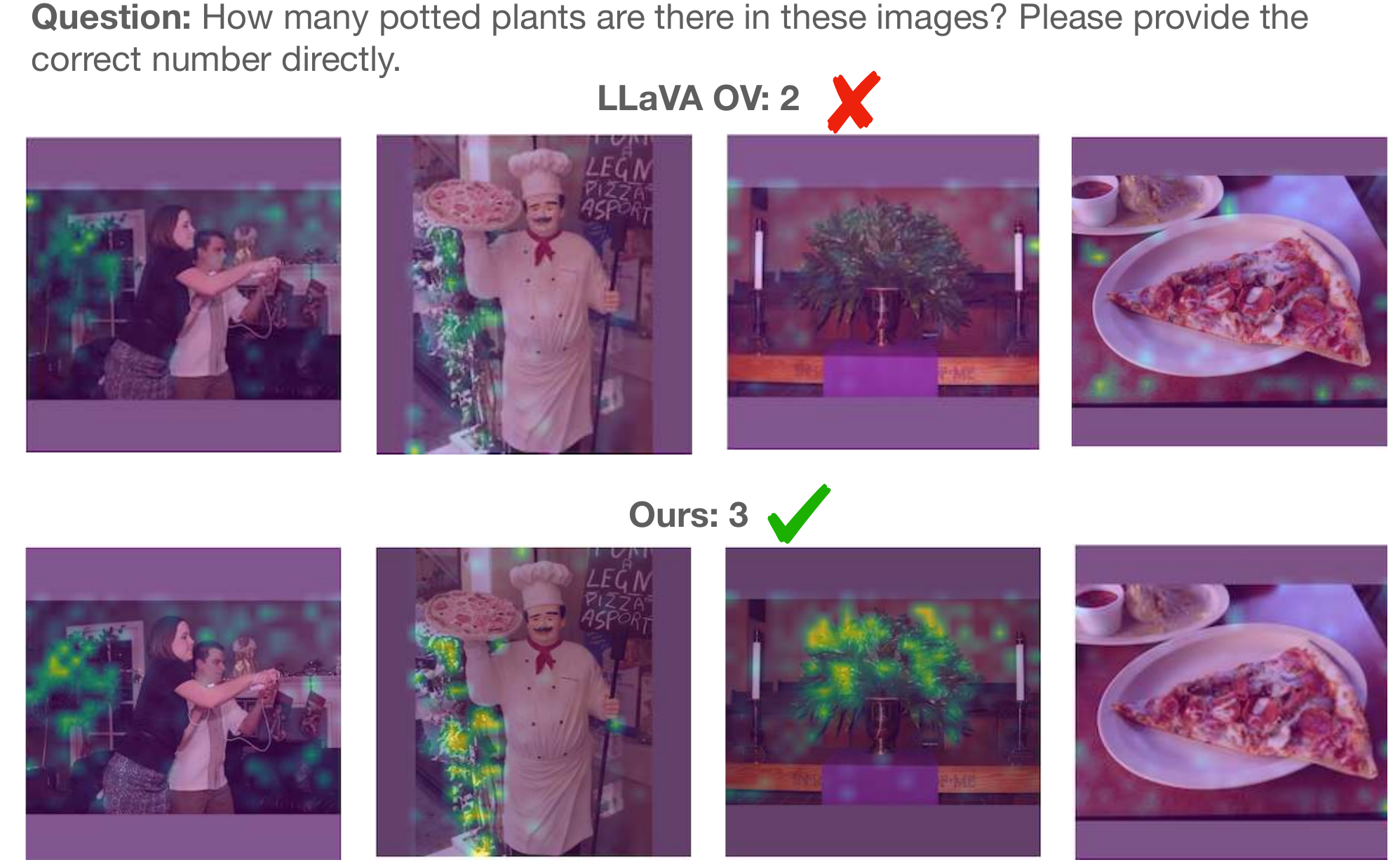}
\caption{\textbf{Answer-to-Image Attention:} The baseline LLaVA OV (top row) fails to attend to the potted plant in the third image, whereas our method (bottom row) correctly focuses on the relevant object. Visualization is shown at the 15th layer of the LLM.}
\label{fig:answer-to-image-attn}
\end{figure}

\myparagraph{Qualitative analysis.} Figure~\ref{fig:answer-to-image-attn} visualizes answer-to-image attention for a `Counting' example. The baseline fails to attend to relevant objects in the third image, resulting in an incorrect count. In contrast, our model exhibits balanced and semantically grounded attention across all images, leading to the correct prediction. This qualitative evidence corroborates our quantitative improvements.

\section{Conclusions}

We systematically investigated the capabilities of LVLMs in multi-image contexts through MIMIC, a novel benchmark designed to isolate specific unitary behaviors. Our analysis reveals that current SOTA models fundamentally exhibit ``single-image behavior,'' struggling to aggregate information across inputs or track multiple concepts in the presence of visual distractors. 
To address this, we introduced a data-centric synthetic fine-tuning strategy and an optimization-centric attention-masking mechanism. These contributions not only resolve key failure modes but also establish new state-of-the-art results, offering a robust foundation for future research in multi-image understanding.

\section{Limitations}

While our work offers a rigorous analysis and effective solutions for multi-image LVLMs, we note the following boundaries of our study:

\begin{itemize}

\item Benchmark Domain: We constructed MIMIC using MS-COCO to maintain precise control over confounding variables (e.g., object counts, occlusion levels). While this design enables exact ``unit testing'' of model reasoning, extending this controlled methodology to specialized domains, such as dense documents or medical imaging, remains an exciting avenue for future research.

\item Resolution Trade-offs: Our analysis demonstrates that reducing sequence length improves multi-image reasoning by mitigating context overload. While highly effective for semantic understanding and counting, tasks requiring pixel-perfect perception of extremely small details might benefit from adaptive resolution strategies, which were outside the scope of this study.

\item Architectural Scope: Our proposed analysis focuses on models with open weights. While we expect conclusions to hold for closed models, additional validations (which induce budget constraints) may be useful for reinforcing our findings.

\end{itemize}

\clearpage

\bibliography{custom}

@string{IJCV = "International Journal on Computer Vision"}

@string{ICCV = "IEEE International Conference on Computer Vision"}

@string{ECCV = "European Conference on Computer Vision"}

@string{CVPR = "IEEE Conference on Computer Vision and Pattern Recognition"}

@string{NeurIPS = "Neural Information Processing Systems"}

@string{ICLR = "International Conference on Learning Representations"}

@string{ICML = "International Conference on Machine Learning"}

@string{WACV = "Winter Conference on Applications of Computer Vision"}

@article{touvron2023llama,
  title={{LLaMA}: Open and efficient foundation language models},
  author={Touvron, Hugo and Lavril, Thibaut and Izacard, Gautier and Martinet, Xavier and Lachaux, Marie-Anne and Lacroix, Timoth{\'e}e and Rozi{\`e}re, Baptiste and Goyal, Naman and Hambro, Eric and Azhar, Faisal and others},
  journal={arXiv preprint arXiv:2302.13971},
  year={2023}
}

@article{abdin2024phi,
  title={Phi-4 technical report},
  author={Abdin, Marah and Aneja, Jyoti and Behl, Harkirat and Bubeck, S{\'e}bastien and Eldan, Ronen and Gunasekar, Suriya and Harrison, Michael and Hewett, Russell J and Javaheripi, Mojan and Kauffmann, Piero and others},
  journal={arXiv preprint arXiv:2412.08905},
  year={2024}
}

@article{jiang2024mixtral,
  title={Mixtral of experts},
  author={Jiang, Albert Q and Sablayrolles, Alexandre and Roux, Antoine and Mensch, Arthur and Savary, Blanche and Bamford, Chris and Chaplot, Devendra Singh and Casas, Diego de las and Hanna, Emma Bou and Bressand, Florian and others},
  journal={arXiv preprint arXiv:2401.04088},
  year={2024}
}

@inproceedings{radford2021learning,
  title={Learning transferable visual models from natural language supervision},
  author={Radford, Alec and Kim, Jong Wook and Hallacy, Chris and Ramesh, Aditya and Goh, Gabriel and Agarwal, Sandhini and Sastry, Girish and Askell, Amanda and Mishkin, Pamela and Clark, Jack and others},
  booktitle=icml,
  year={2021},
}

@inproceedings{zhai2023sigmoid,
  title={Sigmoid loss for language image pre-training},
  author={Zhai, Xiaohua and Mustafa, Basil and Kolesnikov, Alexander and Beyer, Lucas},
  booktitle=iccv,
  year={2023}
}

@inproceedings{fu2024blink,
  title={Blink: Multimodal large language models can see but not perceive},
  author={Fu, Xingyu and Hu, Yushi and Li, Bangzheng and Feng, Yu and Wang, Haoyu and Lin, Xudong and Roth, Dan and Smith, Noah A and Ma, Wei-Chiu and Krishna, Ranjay},
  booktitle=eccv,
  year={2024},
}

@article{wang2024muirbench,
  title={{MuirBench}: A comprehensive benchmark for robust multi-image understanding},
  author={Wang, Fei and Fu, Xingyu and Huang, James Y and Li, Zekun and Liu, Qin and Liu, Xiaogeng and Ma, Mingyu Derek and Xu, Nan and Zhou, Wenxuan and Zhang, Kai and others},
  journal={arXiv preprint arXiv:2406.09411},
  year={2024}
}

@article{openai2023gpt,
  title={Gpt-4 technical report. arxiv 2303.08774},
  author={OpenAI, R},
  journal={View in Article},
  volume={2},
  number={5},
  pages={1},
  year={2023}
}

@article{team2023gemini,
  title={Gemini: a family of highly capable multimodal models},
  author={Team, Gemini and Anil, Rohan and Borgeaud, Sebastian and Alayrac, Jean-Baptiste and Yu, Jiahui and Soricut, Radu and Schalkwyk, Johan and Dai, Andrew M and Hauth, Anja and Millican, Katie and others},
  journal={arXiv preprint arXiv:2312.11805},
  year={2023}
}

@article{jiang2024mantis,
  title={Mantis: Interleaved multi-image instruction tuning},
  author={Jiang, Dongfu and He, Xuan and Zeng, Huaye and Wei, Cong and Ku, Max and Liu, Qian and Chen, Wenhu},
  journal={arXiv preprint arXiv:2405.01483},
  year={2024}
}

@article{laurenccon2023obelics,
  title={Obelics: An open web-scale filtered dataset of interleaved image-text documents},
  author={Lauren{\c{c}}on, Hugo and Saulnier, Lucile and Tronchon, L{\'e}o and Bekman, Stas and Singh, Amanpreet and Lozhkov, Anton and Wang, Thomas and Karamcheti, Siddharth and Rush, Alexander and Kiela, Douwe and others},
  journal=neurips,
  year={2023}
}

@inproceedings{sun2024generative,
  title={Generative multimodal models are in-context learners},
  author={Sun, Quan and Cui, Yufeng and Zhang, Xiaosong and Zhang, Fan and Yu, Qiying and Wang, Yueze and Rao, Yongming and Liu, Jingjing and Huang, Tiejun and Wang, Xinlong},
  booktitle=cvpr,
  year={2024}
}

@misc{liu2024llava,
    title={{LLaVA-NeXT}: Improved reasoning, {OCR}, and world knowledge},
    url={https://llava-vl.github.io/blog/2024-01-30-llava-next/},
    author={Liu, Haotian and Li, Chunyuan and Li, Yuheng and Li, Bo and Zhang, Yuanhan and Shen, Sheng and Lee, Yong Jae},
    month={January},
    year={2024}
}

@inproceedings{liu2024improved,
  title={Improved baselines with visual instruction tuning},
  author={Liu, Haotian and Li, Chunyuan and Li, Yuheng and Lee, Yong Jae},
  booktitle=cvpr,
  year={2024}
}

@article{wang2024cogvlm,
  title={{CogVLM}: Visual expert for pretrained language models},
  author={Wang, Weihan and Lv, Qingsong and Yu, Wenmeng and Hong, Wenyi and Qi, Ji and Wang, Yan and Ji, Junhui and Yang, Zhuoyi and Zhao, Lei and XiXuan, Song and others},
  journal=neurips,
  year={2024}
}

@article{chen2023minigpt,
  title={{MiniGPT}-v2: large language model as a unified interface for vision-language multi-task learning},
  author={Chen, Jun and Zhu, Deyao and Shen, Xiaoqian and Li, Xiang and Liu, Zechun and Zhang, Pengchuan and Krishnamoorthi, Raghuraman and Chandra, Vikas and Xiong, Yunyang and Elhoseiny, Mohamed},
  journal={arXiv preprint arXiv:2310.09478},
  year={2023}
}

@article{li2024llava,
  title={{LLaVA-OneVision}: Easy visual task transfer},
  author={Li, Bo and Zhang, Yuanhan and Guo, Dong and Zhang, Renrui and Li, Feng and Zhang, Hao and Zhang, Kaichen and Zhang, Peiyuan and Li, Yanwei and Liu, Ziwei and others},
  journal={arXiv preprint arXiv:2408.03326},
  year={2024}
}

@article{wang2024qwen2,
  title={Qwen2-vl: Enhancing vision-language model's perception of the world at any resolution},
  author={Wang, Peng and Bai, Shuai and Tan, Sinan and Wang, Shijie and Fan, Zhihao and Bai, Jinze and Chen, Keqin and Liu, Xuejing and Wang, Jialin and Ge, Wenbin and others},
  journal={arXiv preprint arXiv:2409.12191},
  year={2024}
}

@article{yao2024minicpm,
  title={{MiniCPM-V}: A {GPT-4V} level {MLLM} on your phone},
  author={Yao, Yuan and Yu, Tianyu and Zhang, Ao and Wang, Chongyi and Cui, Junbo and Zhu, Hongji and Cai, Tianchi and Li, Haoyu and Zhao, Weilin and He, Zhihui and others},
  journal={arXiv preprint arXiv:2408.01800},
  year={2024}
}

@article{liu2023mitigating,
  title={Mitigating hallucination in large multi-modal models via robust instruction tuning},
  author={Liu, Fuxiao and Lin, Kevin and Li, Linjie and Wang, Jianfeng and Yacoob, Yaser and Wang, Lijuan},
  journal={arXiv preprint arXiv:2306.14565},
  year={2023}
}

@inproceedings{ouali2024clip,
  title={{CLIP-DPO}: Vision-language models as a source of preference for fixing hallucinations in {LVLMs}},
  author={Ouali, Yassine and Bulat, Adrian and Martinez, Brais and Tzimiropoulos, Georgios},
  booktitle=eccv,
  year={2024},
}

@inproceedings{guan2024hallusionbench,
  title={{HallusionBench}: an advanced diagnostic suite for entangled language hallucination and visual illusion in large vision-language models},
  author={Guan, Tianrui and Liu, Fuxiao and Wu, Xiyang and Xian, Ruiqi and Li, Zongxia and Liu, Xiaoyu and Wang, Xijun and Chen, Lichang and Huang, Furong and Yacoob, Yaser and others},
  booktitle=cvpr,
  year={2024}
}

@inproceedings{kaul2024throne,
  title={Throne: An object-based hallucination benchmark for the free-form generations of large vision-language models},
  author={Kaul, Prannay and Li, Zhizhong and Yang, Hao and Dukler, Yonatan and Swaminathan, Ashwin and Taylor, CJ and Soatto, Stefano},
  booktitle=cvpr,
  year={2024}
}

@article{qian2024easy,
  title={How easy is it to fool your multimodal {LLMs}? an empirical analysis on deceptive prompts},
  author={Qian, Yusu and Zhang, Haotian and Yang, Yinfei and Gan, Zhe},
  journal={arXiv preprint arXiv:2402.13220},
  year={2024}
}

@inproceedings{deng2025words,
  title={Words or Vision: Do Vision-Language Models Have Blind Faith in Text?},
  author={Deng, Ailin and Cao, Tri and Chen, Zhirui and Hooi, Bryan},
  booktitle=cvpr,
  year={2025}
}

@article{wang2025text,
  title={Text Speaks Louder than Vision: ASCII Art Reveals Textual Biases in Vision-Language Models},
  author={Wang, Zhaochen and Hooi, Bryan and Wang, Yiwei and Yang, Ming-Hsuan and Huang, Zi and Cai, Yujun},
  journal={arXiv preprint arXiv:2504.01589},
  year={2025}
}

@article{zhang2023video,
  title={Video-{LLaMA}: An instruction-tuned audio-visual language model for video understanding},
  author={Zhang, Hang and Li, Xin and Bing, Lidong},
  journal={arXiv preprint arXiv:2306.02858},
  year={2023}
}

@inproceedings{suhr2019corpus,
  title={A corpus for reasoning about natural language grounded in photographs},
  author={Suhr, Alane and Zhou, Stephanie and Zhang, Ally and Zhang, Iris and Bai, Huajun and Artzi, Yoav},
  booktitle={Proceedings of the 57th annual meeting of the association for computational linguistics},
  pages={6418--6428},
  year={2019}
}

@article{ying2024mmt,
  title={Mmt-bench: A comprehensive multimodal benchmark for evaluating large vision-language models towards multitask agi},
  author={Ying, Kaining and Meng, Fanqing and Wang, Jin and Li, Zhiqian and Lin, Han and Yang, Yue and Zhang, Hao and Zhang, Wenbo and Lin, Yuqi and Liu, Shuo and others},
  journal={arXiv preprint arXiv:2404.16006},
  year={2024}
}

@article{zhao2024benchmarking,
  title={Benchmarking multi-image understanding in vision and language models: Perception, knowledge, reasoning, and multi-hop reasoning},
  author={Zhao, Bingchen and Zong, Yongshuo and Zhang, Letian and Hospedales, Timothy},
  journal={arXiv preprint arXiv:2406.12742},
  year={2024}
}

@article{meng2024mmiu,
  title={Mmiu: Multimodal multi-image understanding for evaluating large vision-language models},
  author={Meng, Fanqing and Wang, Jin and Li, Chuanhao and Lu, Quanfeng and Tian, Hao and Liao, Jiaqi and Zhu, Xizhou and Dai, Jifeng and Qiao, Yu and Luo, Ping and others},
  journal={arXiv preprint arXiv:2408.02718},
  year={2024}
}

@inproceedings{goyal2017making,
  title={Making the {V in VQA} matter: Elevating the role of image understanding in visual question answering},
  author={Goyal, Yash and Khot, Tejas and Summers-Stay, Douglas and Batra, Dhruv and Parikh, Devi},
  booktitle=cvpr,
  year={2017}
}

@inproceedings{mathew2021docvqa,
  title={{DocVQA}: A dataset for vqa on document images},
  author={Mathew, Minesh and Karatzas, Dimosthenis and Jawahar, CV},
  booktitle=wacv,
  year={2021}
}

@inproceedings{kembhavi2016diagram,
  title={A diagram is worth a dozen images},
  author={Kembhavi, Aniruddha and Salvato, Mike and Kolve, Eric and Seo, Minjoon and Hajishirzi, Hannaneh and Farhadi, Ali},
  booktitle=eccv,
  year={2016},
}

@article{stogiannidis2025mind,
  title={Mind the gap: Benchmarking spatial reasoning in vision-language models},
  author={Stogiannidis, Ilias and McDonagh, Steven and Tsaftaris, Sotirios A},
  journal={arXiv preprint arXiv:2503.19707},
  year={2025}
}

@inproceedings{masry2022chartqa,
  title={Chartqa: A benchmark for question answering about charts with visual and logical reasoning},
  author={Masry, Ahmed and Do, Xuan Long and Tan, Jia Qing and Joty, Shafiq and Hoque, Enamul},
  booktitle={Findings of the association for computational linguistics: ACL 2022},
  pages={2263--2279},
  year={2022}
}

@inproceedings{li2024seed,
  title={Seed-bench: Benchmarking multimodal large language models},
  author={Li, Bohao and Ge, Yuying and Ge, Yixiao and Wang, Guangzhi and Wang, Rui and Zhang, Ruimao and Shan, Ying},
  booktitle=cvpr,
  year={2024}
}

@article{fu2024mme,
  title={{MME}-survey: A comprehensive survey on evaluation of multimodal {LLMs}},
  author={Fu, Chaoyou and Zhang, Yi-Fan and Yin, Shukang and Li, Bo and Fang, Xinyu and Zhao, Sirui and Duan, Haodong and Sun, Xing and Liu, Ziwei and Wang, Liang and others},
  journal={arXiv preprint arXiv:2411.15296},
  year={2024}
}

@article{alayrac2022flamingo,
  title={Flamingo: a visual language model for few-shot learning},
  author={Alayrac, Jean-Baptiste and Donahue, Jeff and Luc, Pauline and Miech, Antoine and Barr, Iain and Hasson, Yana and Lenc, Karel and Mensch, Arthur and Millican, Katherine and Reynolds, Malcolm and others},
  journal=neurips,
  year={2022}
}

@article{driess2023palm,
  title={{PaLM-E}: An embodied multimodal language model},
  author={Driess, Danny and Xia, Fei and Sajjadi, Mehdi SM and Lynch, Corey and Chowdhery, Aakanksha and Wahid, Ayzaan and Tompson, Jonathan and Vuong, Quan and Yu, Tianhe and Huang, Wenlong and others},
  journal = ICML,
  year={2023}
}

@article{dai2023instructblip,
  title={{InstructBLIP}: Towards general-purpose vision-language models with instruction tuning},
  author={Dai, Wenliang and Li, Junnan and Li, Dongxu and Tiong, Anthony and Zhao, Junqi and Wang, Weisheng and Li, Boyang and Fung, Pascale N and Hoi, Steven},
  journal=neurips,
  volume={36},
  pages={49250--49267},
  year={2023}
}

@article{zheng2023minigpt,
  title={{MiniGPT-5}: Interleaved vision-and-language generation via generative vokens},
  author={Zheng, Kaizhi and He, Xuehai and Wang, Xin Eric},
  journal={arXiv preprint arXiv:2310.02239},
  year={2023}
}

@article{hong2024cogvlm2,
  title={{CogVLM2}: Visual language models for image and video understanding},
  author={Hong, Wenyi and Wang, Weihan and Ding, Ming and Yu, Wenmeng and Lv, Qingsong and Wang, Yan and Cheng, Yean and Huang, Shiyu and Ji, Junhui and Xue, Zhao and others},
  journal={arXiv preprint arXiv:2408.16500},
  year={2024}
}

@article{zhu2025internvl3,
  title={Internvl3: Exploring advanced training and test-time recipes for open-source multimodal models},
  author={Zhu, Jinguo and Wang, Weiyun and Chen, Zhe and Liu, Zhaoyang and Ye, Shenglong and Gu, Lixin and Tian, Hao and Duan, Yuchen and Su, Weijie and Shao, Jie and others},
  journal={arXiv preprint arXiv:2504.10479},
  year={2025}
}

@inproceedings{chen2024internvl,
    title={Internvl: Scaling up vision foundation models and aligning for generic visual-linguistic tasks},
    author={Chen, Zhe and Wu, Jiannan and Wang, Wenhai and Su, Weijie and Chen, Guo and Xing, Sen and Zhong, Muyan and Zhang, Qinglong and Zhu, Xizhou and Lu, Lewei and others},
    booktitle={Proceedings of the IEEE/CVF Conference on Computer Vision and Pattern Recognition},
    pages={24185--24198},
    year={2024}
  }

@inproceedings{lin2024vila,
  title={Vila: On pre-training for visual language models},
  author={Lin, Ji and Yin, Hongxu and Ping, Wei and Molchanov, Pavlo and Shoeybi, Mohammad and Han, Song},
  booktitle=cvpr,
  year={2024}
}

@article{wang2024longllava,
  title={Longllava: Scaling multi-modal llms to 1000 images efficiently via a hybrid architecture},
  author={Wang, Xidong and Song, Dingjie and Chen, Shunian and Chen, Junyin and Cai, Zhenyang and Zhang, Chen and Sun, Lichao and Wang, Benyou},
  journal={arXiv preprint arXiv:2409.02889},
  year={2024}
}

@inproceedings{hudson2019gqa,
  title={{GQA}: A new dataset for real-world visual reasoning and compositional question answering},
  author={Hudson, Drew A and Manning, Christopher D},
  booktitle=cvpr,
  year={2019}
}

@inproceedings{antol2015vqa,
  title={{VQA}: Visual question answering},
  author={Antol, Stanislaw and Agrawal, Aishwarya and Lu, Jiasen and Mitchell, Margaret and Batra, Dhruv and Zitnick, C Lawrence and Parikh, Devi},
  booktitle=ICCV,
  year={2015}
}

@inproceedings{lin2014microsoft,
  title={Microsoft {COCO}: Common objects in context},
  author={Lin, Tsung-Yi and Maire, Michael and Belongie, Serge and Hays, James and Perona, Pietro and Ramanan, Deva and Doll{\'a}r, Piotr and Zitnick, C Lawrence},
  booktitle=ECCV,
  year={2014},
}

@inproceedings{liu2024mmbench,
  title={Mmbench: Is your multi-modal model an all-around player?},
  author={Liu, Yuan and Duan, Haodong and Zhang, Yuanhan and Li, Bo and Zhang, Songyang and Zhao, Wangbo and Yuan, Yike and Wang, Jiaqi and He, Conghui and Liu, Ziwei and others},
  booktitle=eccv,
  year={2024},
  organization={Springer}
}

@inproceedings{zhao2025mmvu,
  title={Mmvu: Measuring expert-level multi-discipline video understanding},
  author={Zhao, Yilun and Zhang, Haowei and Xie, Lujing and Hu, Tongyan and Gan, Guo and Long, Yitao and Hu, Zhiyuan and Chen, Weiyuan and Li, Chuhan and Xu, Zhijian and others},
  booktitle=cvpr,
  year={2025}
}

@inproceedings{fu2025video,
  title={{Video-MME}: The first-ever comprehensive evaluation benchmark of multi-modal {LLMs} in video analysis},
  author={Fu, Chaoyou and Dai, Yuhan and Luo, Yongdong and Li, Lei and Ren, Shuhuai and Zhang, Renrui and Wang, Zihan and Zhou, Chenyu and Shen, Yunhang and Zhang, Mengdan and others},
  booktitle=cvpr,
  year={2025}
}

@article{wu2024visual,
  title={Visual haystacks: A vision-centric needle-in-a-haystack benchmark},
  author={Wu, Tsung-Han and Biamby, Giscard and Quenum, Jerome and Gupta, Ritwik and Gonzalez, Joseph E and Darrell, Trevor and Chan, David M},
  journal=iclr,
  year={2025}
}

@article{sharma2024losing,
  title={Losing visual needles in image haystacks: Vision language models are easily distracted in short and long contexts},
  author={Sharma, Aditya and Saxon, Michael and Wang, William Yang},
  journal={arXiv preprint arXiv:2406.16851},
  year={2024}
}

@article{chen2024expanding,
  title={Expanding performance boundaries of open-source multimodal models with model, data, and test-time scaling},
  author={Chen, Zhe and Wang, Weiyun and Cao, Yue and Liu, Yangzhou and Gao, Zhangwei and Cui, Erfei and Zhu, Jinguo and Ye, Shenglong and Tian, Hao and Liu, Zhaoyang and others},
  journal={arXiv preprint arXiv:2412.05271},
  year={2024}
}

@article{kuznetsova2020open,
  title={The open images dataset v4: Unified image classification, object detection, and visual relationship detection at scale},
  author={Kuznetsova, Alina and Rom, Hassan and Alldrin, Neil and Uijlings, Jasper and Krasin, Ivan and Pont-Tuset, Jordi and Kamali, Shahab and Popov, Stefan and Malloci, Matteo and Kolesnikov, Alexander and others},
  journal=ijcv,
  volume={128},
  number={7},
  pages={1956--1981},
  year={2020},
}

@inproceedings{ravaut2024context,
  title={On context utilization in summarization with large language models},
  author={Ravaut, Mathieu and Sun, Aixin and Chen, Nancy and Joty, Shafiq},
  booktitle={Annual Meeting of the Association for Computational Linguistics (Volume 1: Long Papers)},
  pages={2764--2781},
  year={2024}
}

\appendix

\clearpage
\section{Appendix}\label{sec:appendix}
\subsection{Additional analysis}

\myparagraph{Multi-image vs single composite image (stitching).} To further understand whether the observed failure modes are due to the presence of multiple images or simply due to the increased number of vision tokens, we conduct an additional experiment involving multi-image stitching: Here, we create a single composite image by stitching multiple images together in a grid format, ensuring that the total number of vision tokens remains similar to that of the multi-image input. The conceptual changes that occur here are twofold: (1) the chat template is devoid of any image separators, and (2) the vision encoder may jointly process parts of different images together. The results across all tasks are summarized in Table~\ref{tab:stiching_experiment}. As the results show, general performance remains similar or increases slightly in some cases.

\begin{figure*}[h]
\centering
\includegraphics[trim=5pt 2pt 5pt 6pt, clip,width=\textwidth]{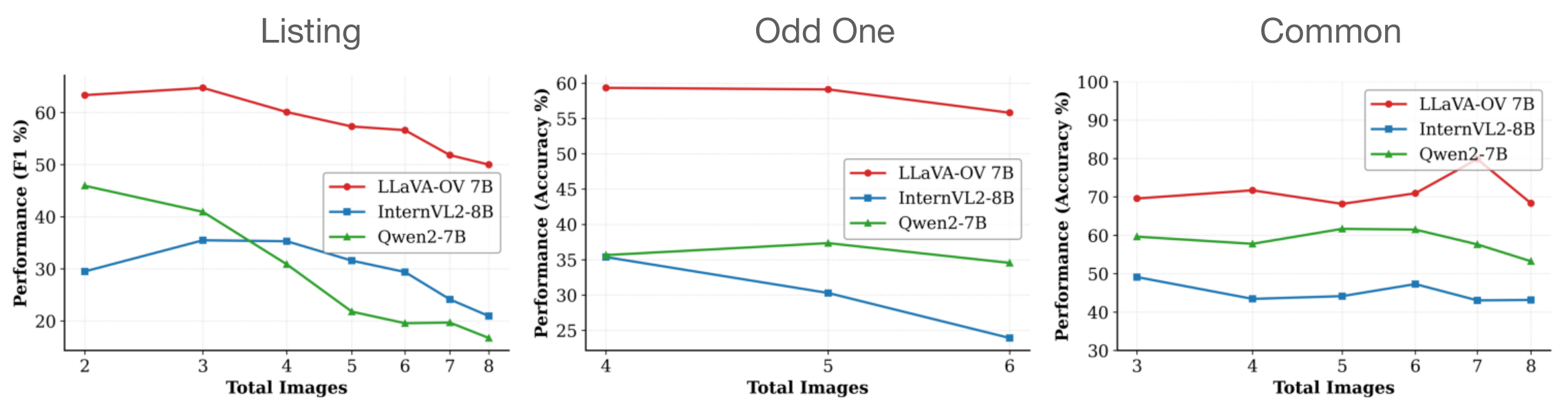}
\caption{\textbf{Performance vs. Number of Images}. 
We report performance as a function of the total number of input images for the Listing, Odd-One tasks, and common tasks.}
\label{fig:query_total_im}
\end{figure*}

\myparagraph{Efficiency and FLOPs Analysis.}
Table~\ref{tab:flops} shows that our masked attention variant achieves superior performance with substantially lower computational cost compared to vanilla attention. On the 0.5B backbone, masked finetuning reduces FLOPs by approximately $81\%$, while also outperforming full finetuning. This demonstrates that selectively constraining inter-image attention is both effective and computationally efficient.

Formally, given $N_t$ text tokens and $N_v$ visual tokens distributed across $M$ independent images, the FLOPs of a standard transformer layer with full self-attention scale as
$\mathcal{O}((N_t + N_v)^2 d + (N_t + N_v) d^2)$,
where $d$ denotes the hidden dimension. The first term corresponds to self-attention computation, while the second accounts for the MLP.

In contrast, our masked attention restricts visual tokens to attend only within their respective image blocks, each of size $n = \frac{N_v}{M}$, while preserving global visibility for text tokens. This modifies the complexity to
\begin{equation}
\text{FLOPs} \approx 
\underbrace{\left( N_t(N_t + N_v) + \sum_{i=1}^{M} n_i^2 \right) d}_{\text{Masked Attention}} 
+ \underbrace{(N_t + N_v)d^2}_{\text{MLP}}.
\end{equation}
Assuming uniform image sizes ($n_i = \frac{N_v}{M}$), this enables scaling to a larger number of high-resolution images under fixed memory and compute budgets.
In our MIMIC benchmark, we observe $M=10.4$ images per sample with an average of $N_t=17.4$ text tokens. Since LLaVA-OV uses 730 visual tokens per image, this yields $N_v=7592$. We use hidden dimensions $d=896$ for the 0.5B model and $d=1536$ for the 1.5B variant, consistent with the FLOPs reductions reported in Table~\ref{tab:flops}.

 {
 \setlength{\tabcolsep}{8pt}
 \renewcommand{\arraystretch}{1.0}
\begin{table}
 \centering
\resizebox{\linewidth}{!}{%
        \begin{tabular}{l|c|c|c|c}
            \toprule
            Model & Common &  Counting &  Odd-one & Listing   \\
            
            \hline 
            LLaVA-OV-0.5B  & 33.6 & 28.9 & 8.8 & 25.0 \\
            LLaVA-OV-0.5B (stitched) & 41.0 & 30.8 & 22.7 & 19.5   \\
            LLaVA-OV-7B &   72.4 & 29.1 & 58.0 & 49.4 \\
            LLaVA-OV-7B (stitched) &  68.2 & 35.9 & 67.1 & 51.5\\
            
            \hline
            Qwen2-VL-2B &  40.4 & 21.8 & 26.3 & 50.8 \\
            Qwen2-VL-2B  (stitched) & 36.2 & 38.8 & 26.8 & 51.1\\
            Qwen2-VL-7B  & 61.3 & 36.6 & 35.3 & 50.3\\
            Qwen2-VL-7B  (stitched) & 51.7 & 35.8 & 51.1 & 58.5 \\

            \hline 
            InternVL2-2B & 25.6 & 28.8 & 7.1 & 35.3  \\
            InternVL2-2B  (stitched) & 29.5 & 29.3 & 6.2 & 39.8 \\
            InternVL2-8B & 42.8 & 30.9 & 24.4 & 52.9 \\
            InternVL2-8B  (stitched) &  42.5 & 27.3 & 37.3 & 54.8 \\ 
                  
            \bottomrule 
         \end{tabular} 
}
\caption{
Stitching Experiment. To ensure a similar number of tokens for both stitched and multi-image, we resize images to 384$\times$384 for LLaVA-OV and 484$\times$484 for Qwen2-VL and InternVL2.}
\vspace{-5pt}
\label{tab:stiching_experiment}
\end{table}
}

{
 \setlength{\tabcolsep}{8pt}
 \renewcommand{\arraystretch}{1.0}
\begin{table}
 \centering

\resizebox{0.8\linewidth}{!}{%
        \begin{tabular}{l|c|c}
            \toprule
             & FLOPs (Gain) &  Avg. Perf.   \\
            \hline
             LLaVA-OV-0.5B  & 58B (0\%) & 26.4\\
            Ours  & 58B (0\%) & 45.5 \\
            \rowcolor{lightgray1} Ours (masked)  & 11.2B (81\%) & 49.4 \\
             \hline 
             LLaVA-OV-1.5B  & 107B (0\%) & 29.8 \\
            Ours  & 107B (0\%) &  54.7\\
             \rowcolor{lightgray1}Ours (masked)  & 26.7B (75\%) & 46.4 \\

            \bottomrule 
         \end{tabular} 
}

\caption{
\textbf{Efficiency analysis.} Our masked-attention variant is substantially more computationally efficient than the vanilla-attention version used in our method.}
\vspace{-5pt}
\label{tab:flops}
\end{table}
}

\myparagraph{Performance vs. number of images.} In Section~\ref{ssec:empirical} and Fig.~\ref{fig:query_total_image}, we analyze how the performance of the counting task varies with the total number of images and the number of query images. Here, we extend this analysis by also considering the Listing, Odd-One, and Common tasks.
Fig.~\ref{fig:query_total_im} reports the performance on these three tasks as the number of input images increases. We observe that for the Listing and Odd-One tasks, performance decreases as the number of total images grows, while performance on the Common task remains largely stable and is not affected by the number of images. This behavior is expected, since the Common task contains no distractors and therefore does not require exhaustive reasoning over all images to predict the correct class.

\begin{figure*}[h]
\centering
\includegraphics[width=\textwidth]{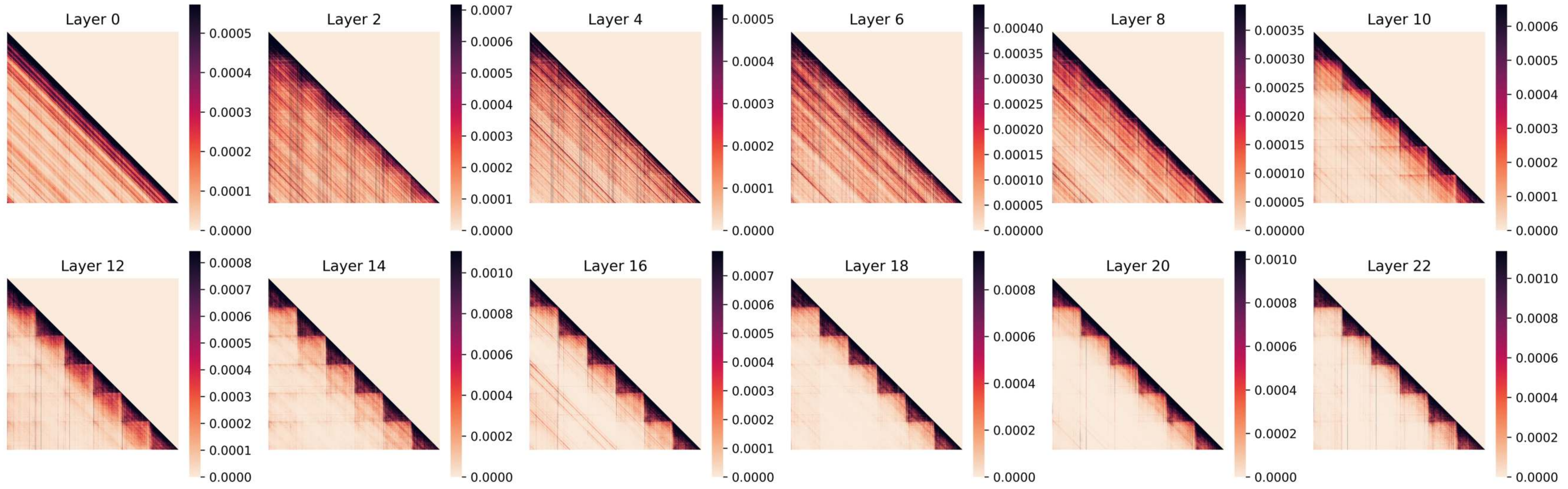}
\caption{\textbf{Inter-image and intra-image token attention across layers for 6 images.}}
\label{fig:inter_intra_image_six}
\end{figure*}

\myparagraph{Extended multi-image interaction.}
Fig.~\ref{fig:im2imattn} in Section~\ref{ssec:empirical} reports the normalized attention scores from each vision token to all other vision tokens in an input sequence of 4 images for a LLaVA-OV model. The scores are computed on a subset of 50 samples and then averaged. Fig.~\ref{fig:inter_intra_image_six} extends this analysis to an input sequence of 6 images. From the figures, we observe that the observations made for 4 images also hold for 6 images. In particular, in the early layers, there is a large amount of inter-image attention, while in deeper layers the attention becomes mostly intra-image, indicating that the model focuses more on individual images. Therefore, we can infer that the observed behavior is intrinsic to the model and does not depend on the number of input images.

\begin{figure}[ht!]
\centering
\includegraphics[width=\linewidth]{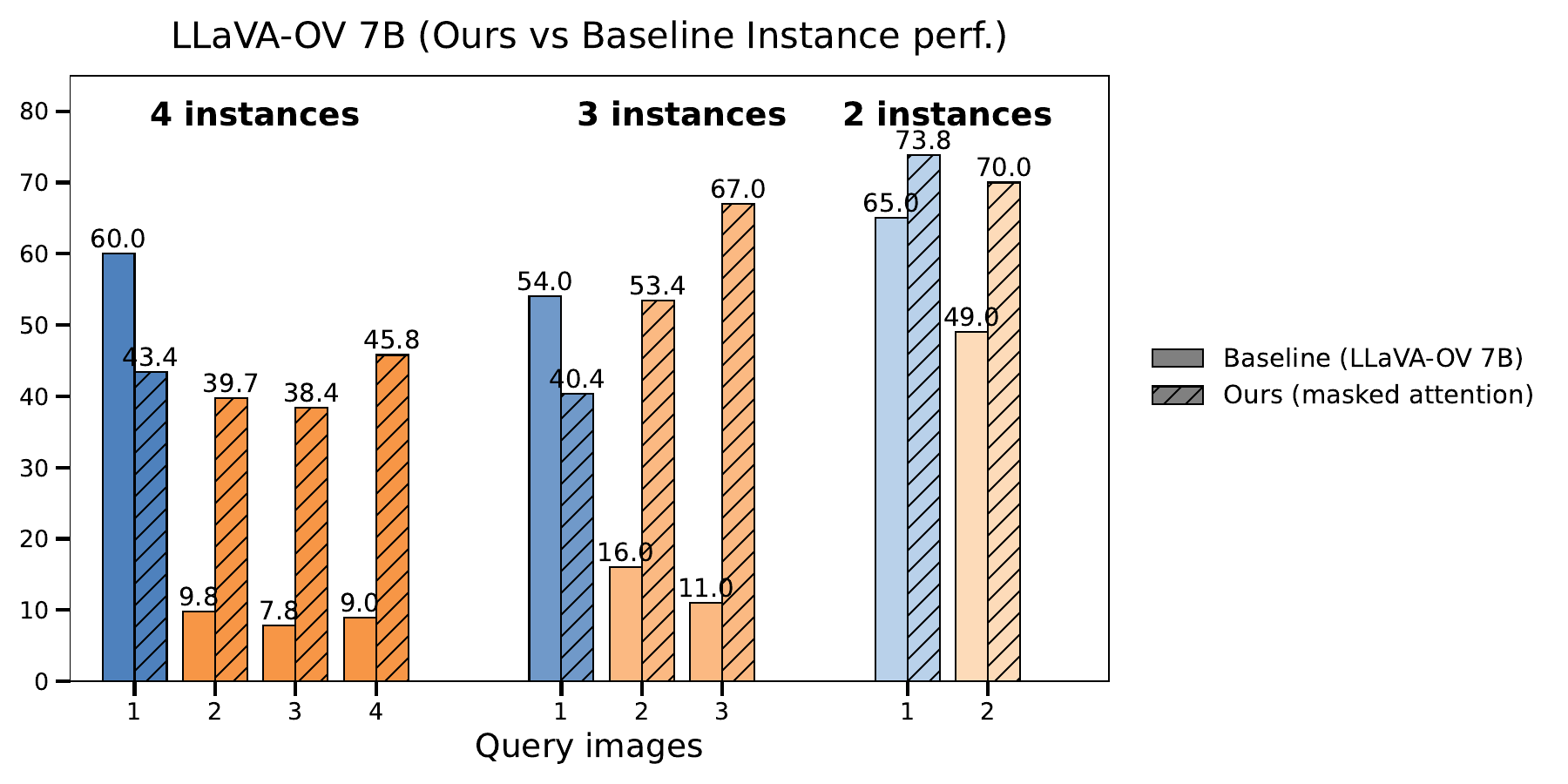}
\caption{\textbf{Comparison on Counting (balanced)}. Our fine-tuned model performs substantially better when object instances are distributed across multiple images. For example, when four instances are spread across four images, performance improves from 9\% to 45.8\%, indicating enhanced cross-image information aggregation.}

\label{fig:before_after_spread}
\end{figure}

\myparagraph{Counting (balanced) Performance Comparison.} We extend the results from~\cref{tab:mimic} with a fine-grained analysis of performance improvements in~\cref{fig:before_after_spread}. We observe that fine-tuning substantially improves performance when object instances are distributed across multiple query images. For example, when four instances are spread across four images, accuracy increases from 9\% to 45.8\%. Similar gains are observed across different instance distributions, indicating improved cross-image information aggregation.

\myparagraph{Extended Performance Comparison with LLaVA-OV 1.5B.} We extend the comparison to multi-image benchmarks from~\cref{tab:mmiu_blink} by including the LLaVA-OV 1.5B model. We observe that both of our fine-tuned models outperform the baseline. In particular, our fine-tuned model achieves a 3.4\% improvement over the baseline, demonstrating enhanced multi-image reasoning capabilities.
{
 \setlength{\tabcolsep}{2pt}
 \renewcommand{\arraystretch}{1.0}
\begin{table}
 \centering
\resizebox{\linewidth}{!}{%
        \begin{tabular}{l |c|c |c | c |c |c |c }
            \toprule
            Model & MuirBench  & Blink & MMIU  &  MIRB & MMT (val) & NLVR2 & Avg. \\
            \hline
            GPT-4V & 62.3 & 54.6 & - & 53.1 & 64.3 & - & - \\
            InternVL2-Llama3-76B~\cite{chen2024expanding} & 51.2 & 56.8 & 44.2 & 58.2 & 67.4 & - & -\\
            LLaVA-v1.5-7B & 20.0 & 37.1 & 19.2 & 28.5 & - & - & - \\  
            InternVL2-2B~\cite{chen2024expanding}  & 24.3 & 16.3 & 13.6 & 25.0 & 46.7 & 18.9 & 24.1 \\
            InternVL2-8B~\cite{chen2024expanding}  & 37.9 & 23.4 & 36.8 & 48.6 & 57.9 & 8.7 & 35.6 \\
            Qwen2VL-2B~\cite{wang2024qwen2} & 27.2 & 12.7 & 38.7 & 45.9 &  51.9 & 41.6 & 36.3 \\
            Qwen2VL-7B~\cite{wang2024qwen2} & 43.0 & 17.7 & 52.6 & 60.8 & 61.7 & 41.5 & 46.2 \\ 
            \hline
            
             LLaVA-OV-0.5B~\cite{li2024llava} & 26.8  & 40.4 & 34.2 & 31.8 & 41.1 & 61.2 & 39.3 \\
       
           \rowcolor{lightgray1}  Ours & 33.6  & 38.9 & 37.2 & 32.8 & 45.6 & 68.0 & 42.7\\
       
            \rowcolor{lightgray1} Ours (masked) & 32.5 & 39.1 & 36.3 & 28.5 & 45.9 & 65.1 & 41.2 \\

             \hline 

             LLaVA-OV-1.5B~\cite{li2024llava} & 31.1 &  36.4 & 33.4 & 37.7 & 47.5 & 70.9 & 42.8  \\
           \rowcolor{lightgray1}  Ours & 39.7 & 40.0 & 38.9 & 36.0 & 48.8 & 73.7 & 46.2 \\
           \rowcolor{lightgray1}  Ours (masked) & 32.3 & 42.2 & 35.5 & 30.6 & 48.1 & 69.0 & 43.0  \\
             \hline 
             LLaVA-OV-7B~\cite{li2024llava} & 41.7  & 50.4 & 45.0 & 47.2 & 56.6 & 84.2 & 54.2 \\

            \rowcolor{lightgray1} Ours (masked) & 51.3 & 51.9 & 45.5 & 51.0 & 55.3 & 87.3 & 57.1\\
            \bottomrule 
         \end{tabular} 
}
\caption{
Extended Comparisons including LLaVA-OV 1.5B with the state-of-the-art on multi-image benchmarks: MuirBench, Blink, MMIU, MIRB, MMT and NLVR2. 
}
\vspace{-5pt}
\label{tab:mmiu_blink_15b}
\end{table}
}

\begin{figure}[ht!]
\centering
\includegraphics[width=0.8\linewidth]{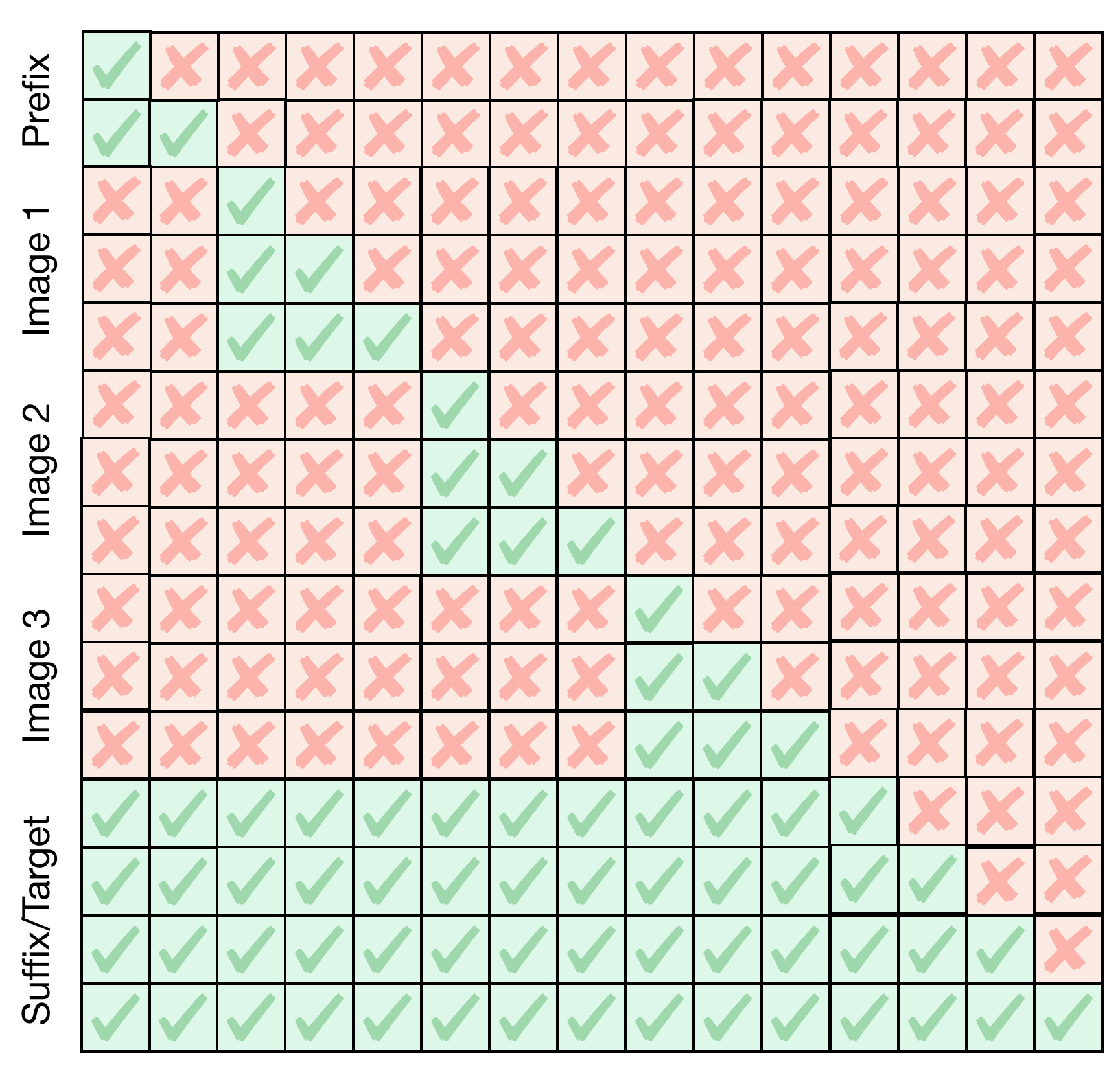}
\caption{\textbf{Our Masked Attention.} Vision tokens are restricted to attend only to tokens from the same image, following a block-diagonal attention pattern, while text tokens in both the prefix and suffix follow the standard autoregressive attention.}
\label{fig:masked_attn}
\end{figure}

\begin{figure*}[ht!]
\centering
\includegraphics[trim=10pt 2pt 5pt 6pt, clip,width=0.8\linewidth]{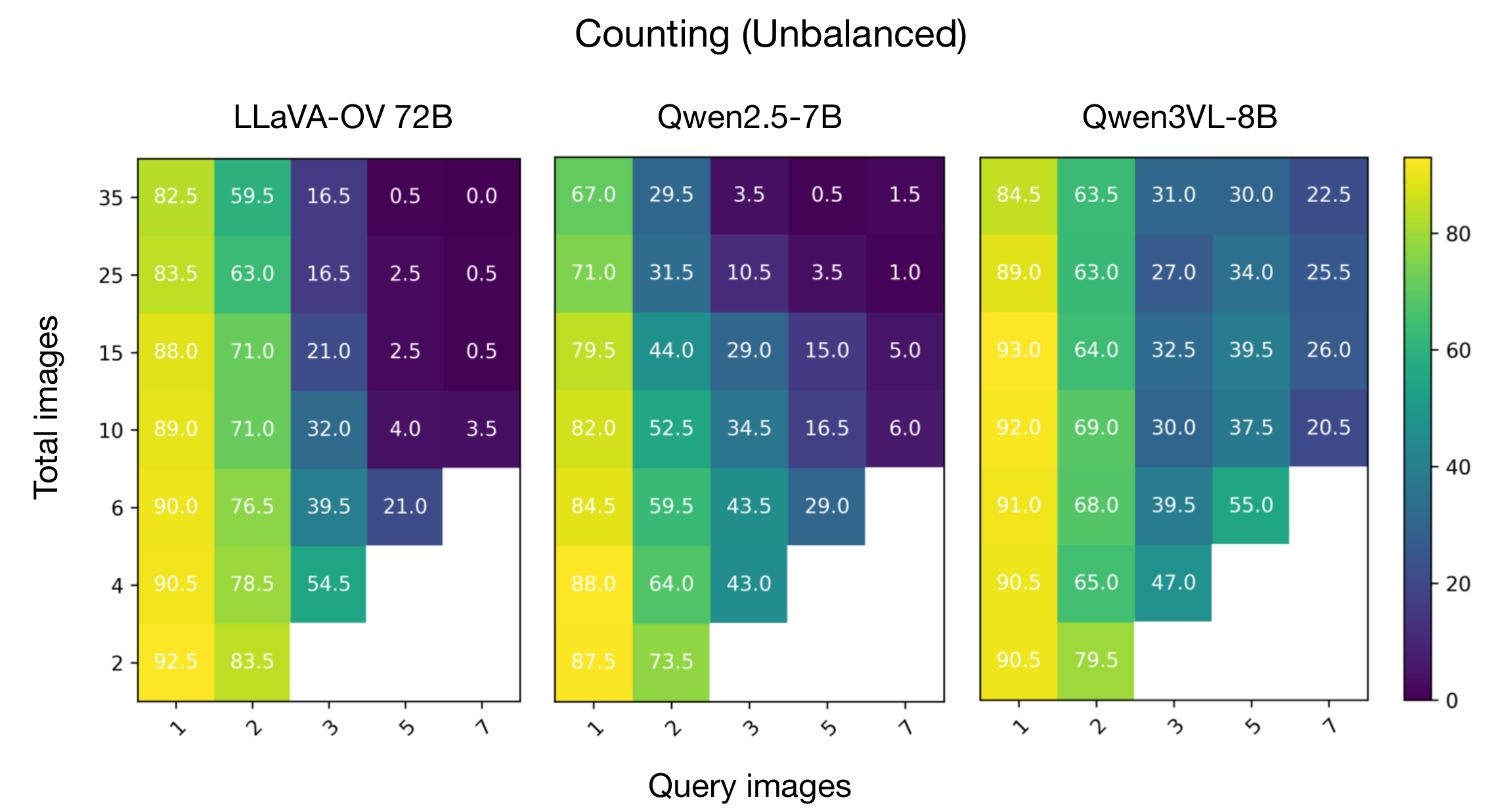}
\caption{
\textbf{Counting (unbalanced) performance with bigger and latest models}. We analyze the trade-off between the number of query images and the total number of images for bigger (LLaVA-OV 72B) and latest models (Qwen2.5-7B and Qwen3VL-8B).}
\label{fig:query_total_image_big}
\end{figure*}

\myparagraph{Bigger and latest models.}
Fig.~\ref{fig:query_total_image_big} reports results on the Counting task by increasing the number of images that contain the object instance from 1 to 7. We consider larger (LLaVA-OV 72B) or more recent (Qwen2.5-7B, Qwen3-VL-8B) models compared to those analyzed in Fig.~\ref{fig:query_total_image} (left). Consistent with previous findings outlined in Section~\ref{ssec:empirical}, even more powerful models achieve strong performance when the object of interest appears in a single query image, but performance decreases as the same object is spread across multiple images.

\begin{figure*}[ht!]
\centering
\includegraphics[width=\linewidth]{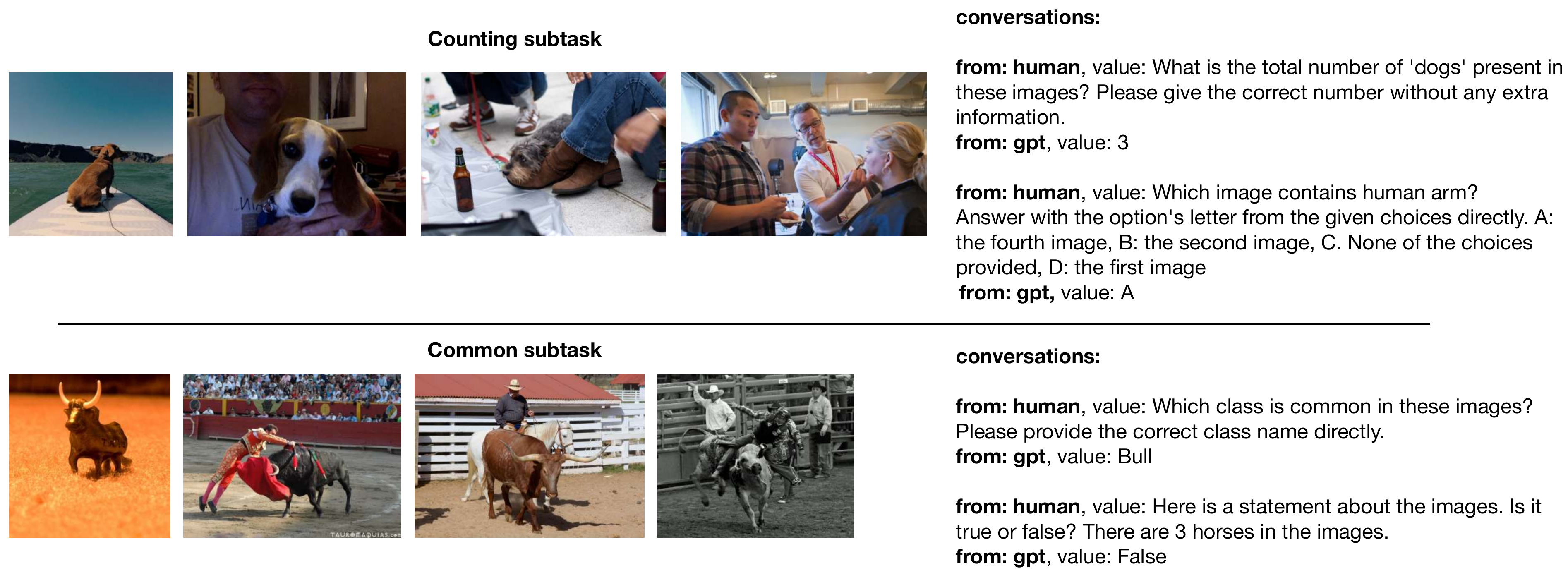}
\caption{
\textbf{Samples from the MIMIC training dataset}.Unlike the evaluation benchmark, the training data follows the LLaVA data format and includes multi-turn conversations with option-based answers.}
\label{fig:training_sample}
\end{figure*}

\subsection{Implementation Details.}

\myparagraph{Training Data.}
Our method is trained on a unified training set composed of synthetic samples generated using the MIMIC pipeline (see Section 3.1) together with the original LLaVA-OV multi-image instruction-tuning data (approximately 580K samples). Unlike the MIMIC benchmark used for evaluation, the synthetic MIMIC training dataset is built from OpenImages~\cite{kuznetsova2020open} annotations and provides explicit supervision for cross-image reasoning. Additionally, it supports multi-turn conversations and option-based responses (see ~\cref{fig:training_sample}). Dataset statistics are reported in~\cref{tab:training_statistics}. In particular, the dataset consists of approximately 50K samples from each MIMIC subtask, with sequences containing up to 10 images. This design encourages learning effective cross-image information aggregation while retaining general vision-language capabilities through joint training with LLaVA-OV data.

\myparagraph{Training Procedure.}
For both training strategies—the data-centric fine-tuning approach and the optimization-centric attention-masking approach—we start from the LLaVA-OV Single-Image variant (Stage 2.1)~\cite{li2024llava}, which has been pre-trained on high-quality single-image instruction-tuning data. We freeze the vision encoder and train only the language model layers and the vision-to-language projector.

For the optimization-centric attention-masking strategy, we do not fully fine-tune the language model layers; instead, we apply LoRA adapters to these layers. In addition, we introduce attention masking in the self-attention blocks, restricting vision tokens to attend only to tokens from the same image. For the data-centric fine-tuning strategy, we fully fine-tune the language model layers without attention masking.

All models are trained with an effective batch size of 128, zero weight decay, and a cosine learning rate schedule with a warmup ratio of 0.03. For the data-centric approach, we use a learning rate of $2.5\times1e^{-6}$, while for the LoRA-based attention-masking strategy, we increase the learning rate to $2.5\times1e^{-5}$. Training is performed on 8 NVIDIA H100 GPUs with approximately 80GB memory each. For the attention-masking strategy, we set the LoRA rank to 128.

\subsection{Additional MIMIC details}

{
 \setlength{\tabcolsep}{4pt}
 \renewcommand{\arraystretch}{1.0}
\begin{table}[ht!]
 \centering
\resizebox{\linewidth}{!}{%
        \begin{tabular}{l c c c c c }
            \toprule
             
            & Counting & Common & Odd One & Listing & Overall \\
            \midrule
            Queries  &  50000 & 50000 & 47561 & 49267 & 196828 \\
         
            Images &  157995 & 80438 & 81642 & 108300 & 232196\\
            Obj. inst. per query  & 27.3 & 16.3 & 17.8 & 18.6 & 20.0 \\
            Min img per query  & 2 & 2 & 3 & 2 & 2 \\
            Max img per query & 10 & 8 & 8 & 8 & 10 \\
            Median img per query  & 5 & 4 & 4 & 4 & 4  \\
            Avg words per question & 48.9 & 44.6 & 47.7 & 52.1 & 48.3 \\
            \bottomrule 
         \end{tabular} 
}
\caption{MIMIC synthetic training dataset statistics based on OpenImagesv7~\cite{kuznetsova2020open}.}
\vspace{-5pt}
\label{tab:training_statistics}
\end{table}
}

\begin{figure*}[h]
\centering
\includegraphics[width=\textwidth]{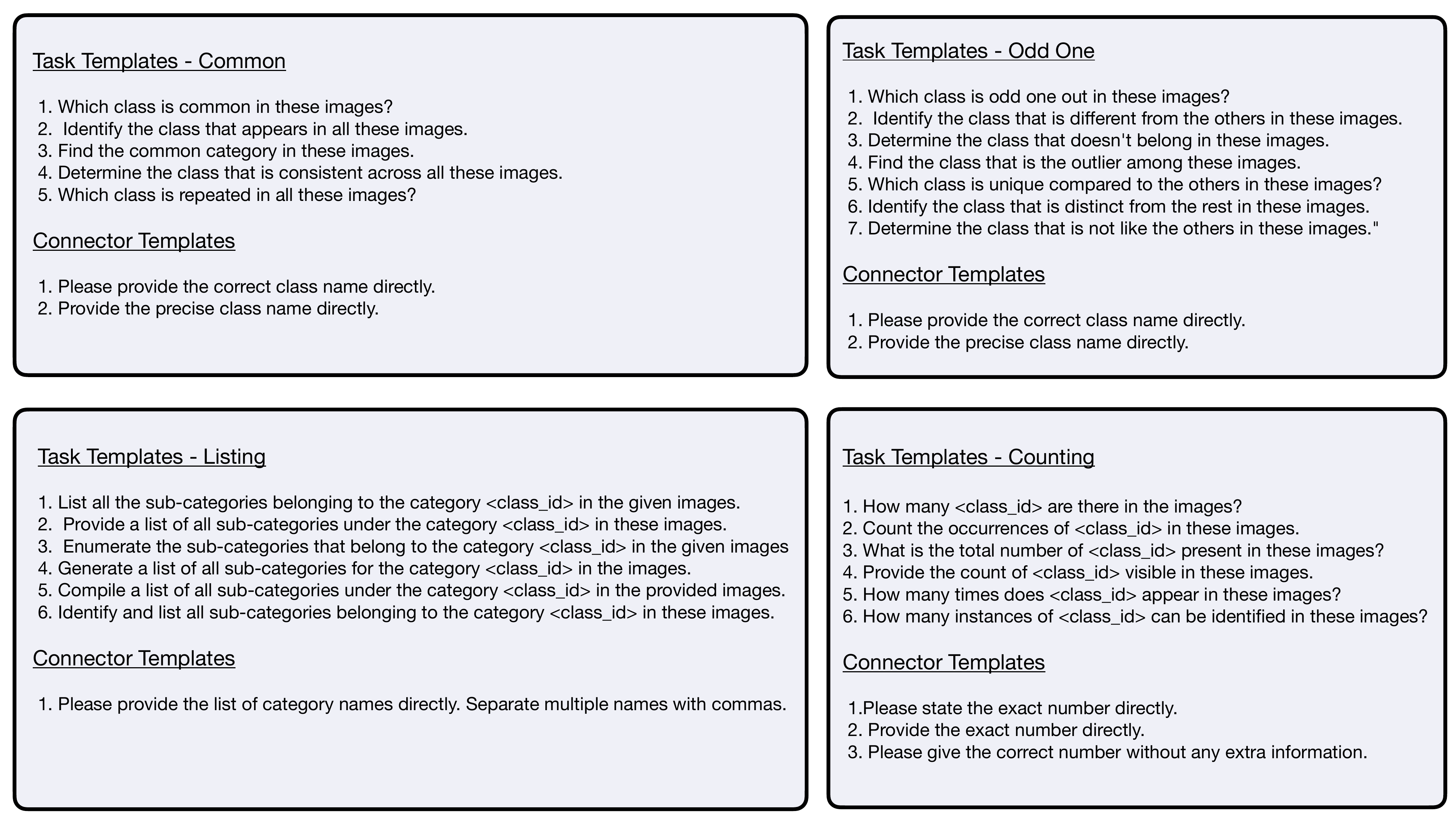}
\caption{\textbf{Prompt Templates for Different Tasks.} For each task, a prompt is constructed by randomly sampling one template from the task-specific template set ($P_{\text{task}}$) and one from the connector template set ($P_{\text{connector}}$). The two templates are then combined to form the final prompt, i.e., $P = P_{\text{task}}\Vert P_{\text{connector}}$. }
\label{fig:prompt_templates}
\end{figure*}

\myparagraph{Prompt templates}
Fig.~\ref{fig:prompt_templates} shows the different prompt templates used for the four tasks.
For each task, a prompt is constructed by randomly sampling one template from the task-specific template set ($P_{\text{task}}$) and one from the connector template set ($P_{\text{connector}}$). The two templates are then combined to form the final prompt, i.e., $P = P_{\text{task}}\Vert P_{\text{connector}}$.

\begin{figure*}[t]
\centering
\includegraphics[width=\textwidth]{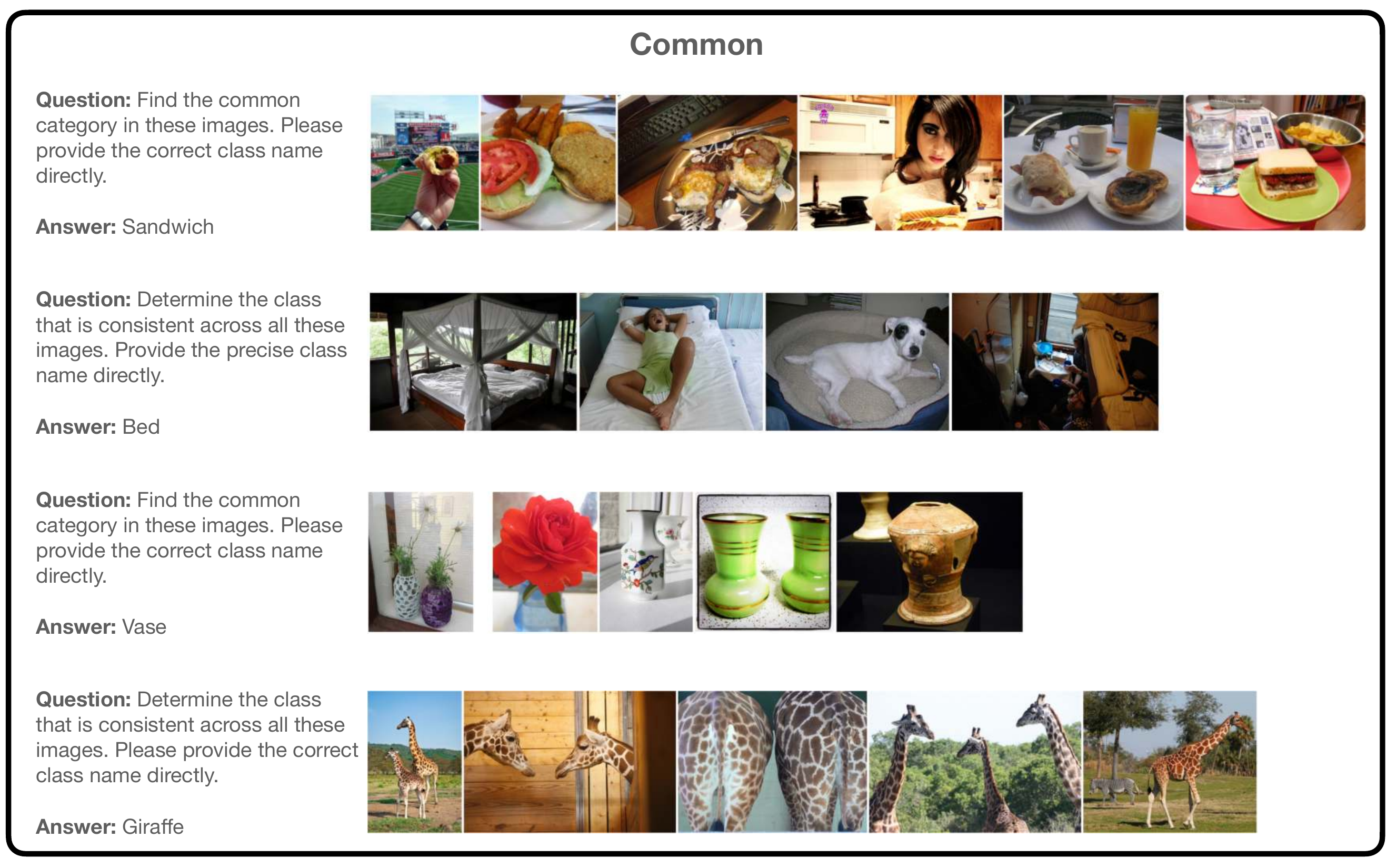}
\caption{Our MIMIC Evaluation Benchmark. We show some samples from the category `common' from our benchmark.}
\label{fig:common_ext}
\end{figure*}

\begin{figure*}[t]
\centering
\includegraphics[width=\textwidth]{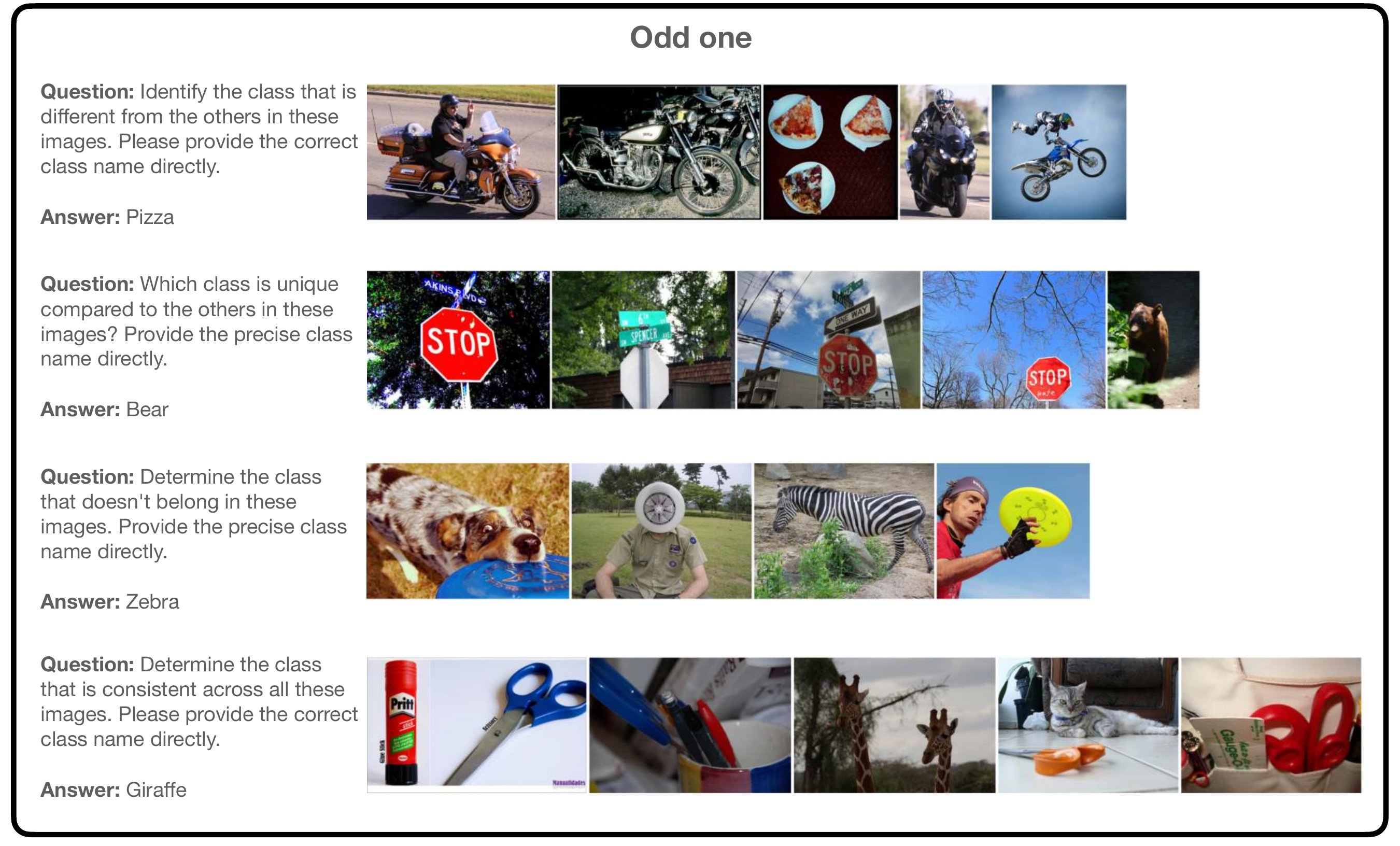}
\caption{Our MIMIC Evaluation Benchmark. We show some samples from the category `Odd One' from our benchmark.}
\label{fig:oddone_ext}
\end{figure*}

\begin{figure*}[t]
\centering
\includegraphics[width=\textwidth]{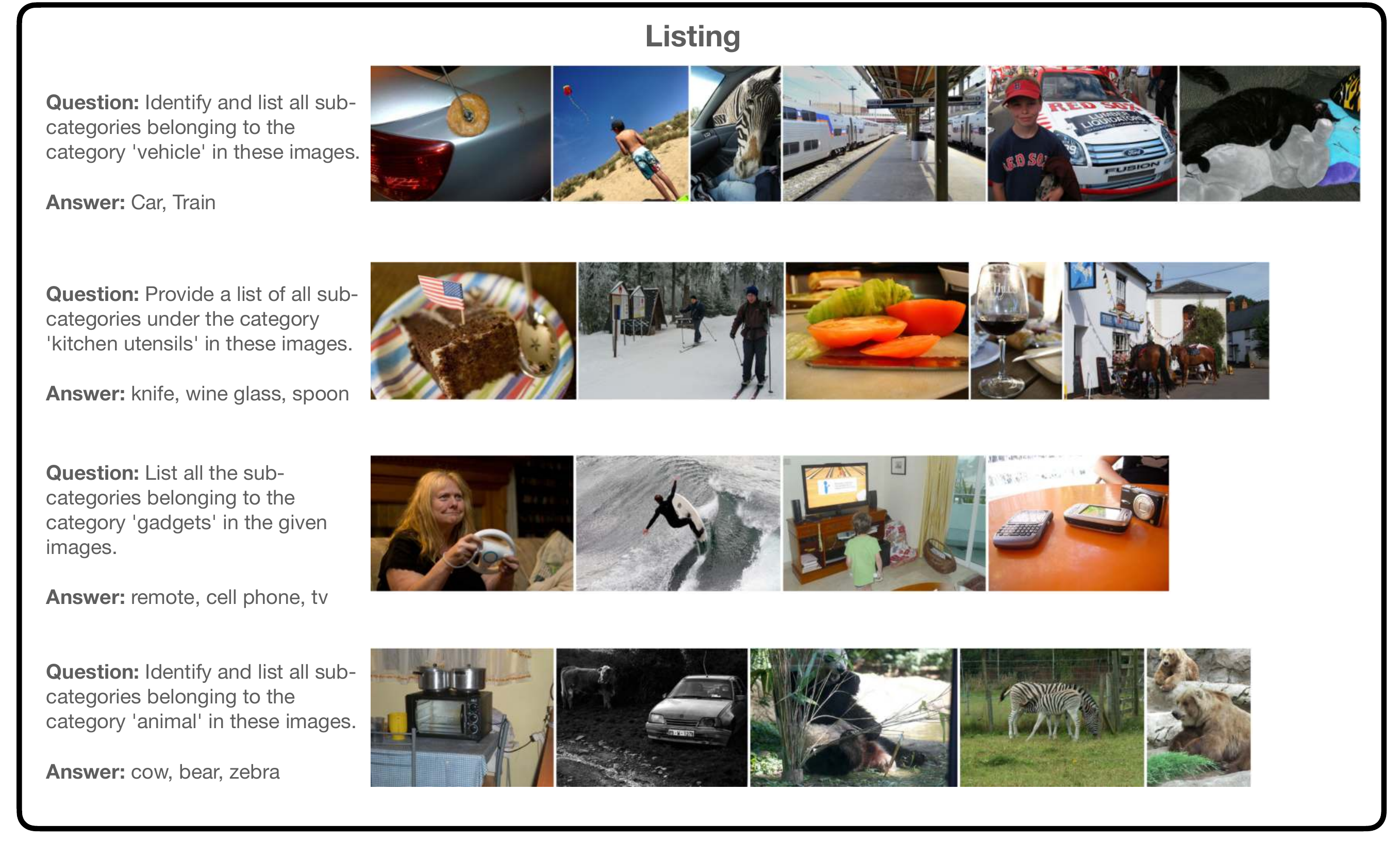}
\caption{Our MIMIC Evaluation Benchmark. We show some samples from the category `Listing' from our benchmark.}
\label{fig:listing_ext}
\end{figure*}

\begin{figure*}[!t]
\includegraphics[width=\textwidth]{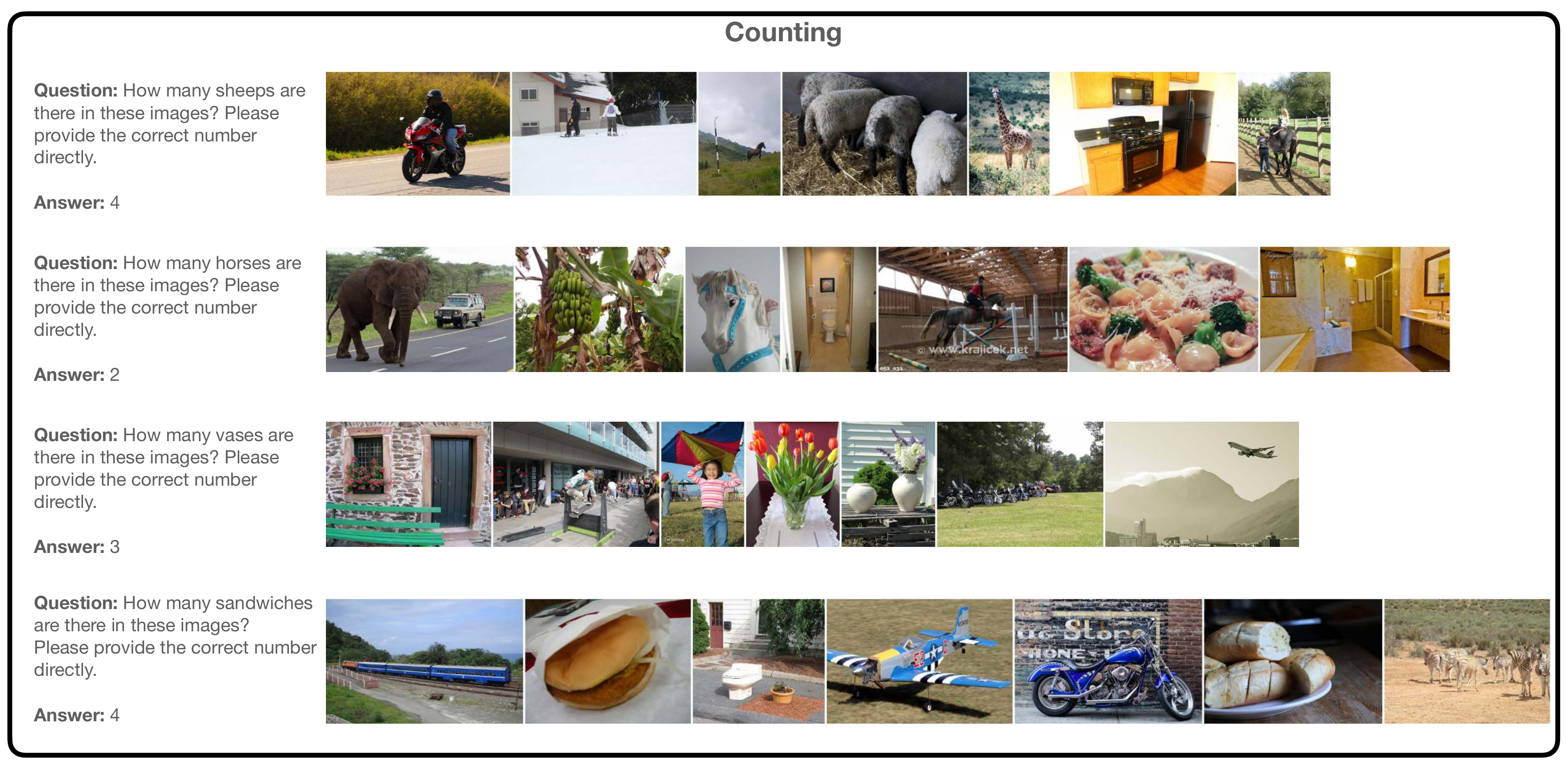}
\caption{Our MIMIC Evaluation Benchmark. We show some samples from the category `Counting' from our benchmark.}
\label{fig:counting_ext}
\vspace{14cm}
\end{figure*}

\myparagraph{Additional examples}
We report additional sample instances from the Common, Odd-One, Listing, and Counting categories, in Figs.~\ref{fig:common_ext},~\ref{fig:oddone_ext},~\ref{fig:listing_ext} and ~\ref{fig:counting_ext}, respectively.

\clearpage

\end{document}